%% file: neurips_2026_arxiv.tex
\documentclass[numbers]{article}
\usepackage{float}
\usepackage{graphicx}
\usepackage[main, final]{neurips_2026}

\usepackage[utf8]{inputenc}
\usepackage[T1]{fontenc}
\usepackage{hyperref}
\usepackage{url}
\usepackage{booktabs}
\usepackage{amsfonts}
\usepackage{nicefrac}
\usepackage{microtype}
\usepackage[table]{xcolor}
\usepackage{amsmath}
\usepackage{amssymb}
\usepackage{multirow}
\usepackage{siunitx}
\usepackage{float}
\setcitestyle{numbers}
\usepackage{makecell}
\usepackage{rotating}

\newcommand{\NAME}{{HRF-Res}}
\newcommand{\LIFRC}{{LIF-Res}}

\title{Beyond Low-Pass Dynamics: Frequency-Selective Spiking Reservoirs with Resonant Neurons}

\author{%
  Aurora Micheli \\
  Delft University of Technology \\
  \texttt{a.micheli@tudelft.nl} \\
  \And
  Claudio Gallicchio \\
  University of Pisa \\
  \texttt{claudio.gallicchio@unipi.it} \\
  \And
  Nergis Tomen \\
  Delft University of Technology \\
  \texttt{n.tomen@tudelft.nl} \\
}

\begin{document}
\maketitle

\begin{abstract}
Spiking reservoir computing combines the training efficiency of fixed recurrent reservoirs with the energy efficiency of event-driven inference, where computation is triggered by sparse spikes rather than executed densely at every timestep. However, existing spiking reservoirs almost exclusively rely on Leaky Integrate-and-Fire (LIF) neurons, whose low-pass dynamics limit their ability to distinguish information encoded at different temporal frequencies. We introduce \NAME, a liquid state machine in which LIF neurons are replaced by a heterogeneous population of Harmonic Resonate-and-Fire (HRF) neurons. These neurons exhibit intrinsic oscillatory dynamics and selectively respond to input frequencies, enabling the reservoir to capture frequency-specific temporal patterns. The result is more discriminative representations and higher classification accuracy than LIF-based reservoirs, achieved without sacrificing the energy efficiency of spiking computation, with up to 19$\times$ lower energy consumption than an equivalent non-spiking oscillatory reservoir.
Evaluated on ten benchmarks spanning time series, sequential images, audio spike trains, and neuromorphic vision, \NAME~achieves state-of-the-art accuracy among spiking reservoir methods.
These results demonstrate that introducing frequency-selective dynamics into spiking reservoirs improves representational quality and classification performance while preserving their inherent energy efficiency. 

\end{abstract}

\section{Introduction}
\label{sec:intro}

Processing temporal sensor streams at the edge, from industrial vibration
monitoring to event-based vision and neuromorphic audio, requires models that
are both computationally efficient and capable of operating under strict power
constraints. Modern sequence models such as
LSTMs~\cite{hochreiter1997long}, GRUs~\cite{cho2014learning}, and
Transformers~\cite{vaswani2017attention} achieve strong performance but were
not designed for this regime: they require backpropagation through time (BPTT)
to train, and perform dense computation at every timestep during inference. As a result, their energy consumption scales with model size rather than input activity, making them inefficient for
low-power, event-driven settings where signals are sparse and computation should ideally be triggered only when new information arrives.

Reservoir Computing (RC) addresses the training cost: a fixed, randomly
initialised recurrent network projects inputs into a rich dynamical state and
only a linear readout is trained~\cite{jaeger2001echo,maass2002real},
eliminating BPTT entirely. The inference cost, however, remains: RC reservoirs
perform dense computation at every timestep, so energy still scales with
network size rather than input activity. Spiking Neural Networks (SNNs)
offer a complementary solution. By encoding information as discrete spike
events, computation is triggered only by activity, making energy consumption
proportional to firing rate rather than network size~\cite{maass1997networks,
paugam2012computing}. This event-driven paradigm maps naturally onto
neuromorphic hardware such as Intel Loihi~\cite{davies2018loihi}, IBM
TrueNorth~\cite{merolla2014truenorth}, and SpiNNaker~\cite{furber2014spinnaker},
and is particularly well suited to event-based sensing
modalities~\cite{gallego2022eventsurvey}. Combining SNNs with RC yields
Liquid State Machines (LSMs)~\cite{maass2002real}, which inherit both
training efficiency and activity-driven inference. Recent work has shown
competitive LSM performance on temporal benchmarks~\cite{deckers2022elsm,
dey2022efficient,biswas2024icons}.

Despite this promise, existing spiking reservoirs almost exclusively rely on leaky integrate-and-fire (LIF) neurons. A LIF neuron acts as a temporal integrator~\cite{burkitt2006review}: inputs that vary rapidly over time are increasingly attenuated, resulting in an effectively low-pass response. A reservoir composed of such units therefore encodes how much input arrives,
but not at which frequencies. For signals with multiple temporal scales, this constitutes a fundamental representational limitation~\cite{whichmodel}.

Oscillatory dynamics provide a natural mechanism to overcome this lack of sensitivity to temporal frequency structure.
Subthreshold membrane oscillations have been widely observed in biological
neurons~\cite{llinas1988intrinsic,alonso1989subthreshold} and are believed to
play a key role in temporal coding~\cite{buzsaki2004neuronal,effenberger2025functional}. In artificial
systems, networks of oscillators form a powerful computational substrate for
temporal processing~\cite{lanthaler2023neural}, and non-spiking oscillatory
reservoirs such as RON~\cite{ceni2024ron} show that heterogeneous resonance
frequencies can improve performance and stability. Whether these benefits
can be realised within a spiking reservoir remains an open question.

In this work, we instantiate oscillatory dynamics in a spiking setting using Resonate-and-Fire neurons, which incorporate subthreshold oscillatory behaviour within a spike-generating mechanism~\cite{izhikevich2001resonate,AlKhamissi2021}. Specifically, we employ Harmonic Resonate-and-Fire (HRF) neurons, which exhibit damped oscillatory dynamics and intrinsic resonance frequencies. Unlike LIF neurons, HRF neurons respond preferentially to inputs matching their resonance, effectively acting as band-pass filters. Intuitively, a heterogeneous population of HRF neurons behaves like a bank of tuning forks, each responding to its own frequency band, so that the reservoir state reflects not only how much input arrives but also at which temporal frequencies. By replacing LIF units with a heterogeneous population of HRF neurons, we introduce \NAME, a Liquid State Machine that combines the efficiency of spiking computation with frequency-selective representations.

Our aim is not the neuron substitution per se, but to establish frequency selectivity as a computational principle for spiking reservoirs. Through a controlled comparison in which neuron dynamics are the only difference between models, we test whether band-pass behaviour survives in a full recurrent spiking network, whether it makes reservoir representations more class-discriminative, and whether this translates into higher accuracy at low energy cost.
Our contributions are as follows:
\begin{itemize}
    \item We show that HRF neurons act as distributed band-pass filters, producing 
    intrinsically more class-discriminative representations than an equivalent LIF 
    reservoir before any readout is trained. A controlled ablation confirms that this representational advantage translates into improved classification
    accuracy across multiple datasets and modalities.
    \item \NAME~achieves up to 19$\times$ lower energy
    consumption compared to the non-spiking RON baseline~\cite{ceni2024ron},
    with only modest accuracy loss.
    \item We evaluate the proposed model on ten benchmarks spanning
    univariate time series, sequential images, audio spike trains, and
    neuromorphic vision, achieving state-of-the-art performance among spiking reservoir methods.
\end{itemize}
Code to reproduce all experiments and results reported in this paper is available at \url{https://github.com/AuroraMicheli/HRF-Res}.

\section{Related Work}
\label{sec:related}

\textbf{Reservoir computing and oscillatory dynamics.}
Reservoir Computing (RC)~\cite{jaeger2001echo,maass2002real} provides an
efficient framework for temporal processing by decoupling representation
learning from training. Its dynamical properties and practical usage have been
extensively studied~\cite{lukosevicius2009reservoir,lukosevicius2012practical}.
A key factor in RC performance is the choice of reservoir dynamics. Early work
on Liquid State Machines already suggested that neuron models with resonant or oscillatory subthreshold dynamics can yield more separable representations than standard integrator
neurons~\cite{whichmodel}. More recently, oscillatory systems have
been revisited as a powerful computational substrate, both from a theoretical
perspective~\cite{lanthaler2023neural} and in practice: Ceni et
al.~\cite{ceni2024ron} showed that reservoirs composed of heterogeneous damped
oscillators significantly outperform standard RC models. However, this approach relies on continuous-valued units, performing dense computation at every timestep, and therefore does not benefit from the energy savings of event-driven inference.
Beyond reservoir computing, oscillatory dynamics have recently been explored as a general inductive bias in gradient-trained deep networks, from coupled-oscillator RNNs and GNNs~\cite{rusch2021coupled,rusch2021unicornn,rusch2022graphcon} to Kuramoto-oscillator units in modern architectures~\cite{miyato2025akorn}, building on earlier work on neural synchrony via complex-valued units~\cite{reichert2014neuronal}.

\textbf{Spiking reservoir computing.}
Liquid State Machines (LSMs)~\cite{maass2002real} extend RC to spiking neural
networks, enabling event-driven computation and compatibility with
neuromorphic hardware. A large body of work has explored architectural and
parametrisation improvements, including heterogeneous time
constants~\cite{deckers2022elsm}, efficient implementations and scaling
strategies~\cite{dey2022efficient,patino2022spinnaker}, and ensemble or
receptive-field approaches~\cite{biswas2024icons}. Some recent works also
consider hybrid models that depart from the strict RC paradigm, such as
Gaurav et al.~\cite{gaurav2023reservoir}, which combine non-spiking feature
extractors with trainable spiking components. Despite these advances, existing
spiking reservoirs predominantly rely on leaky integrate-and-fire (LIF)
neurons, which lack intrinsic frequency selectivity.

\textbf{Resonate-and-fire neurons.}
Resonate-and-fire (RF) neurons~\cite{izhikevich2001resonate} introduce
oscillatory membrane dynamics into spiking models, enabling selective responses to input frequency.  Their computational properties have been explored in a
variety of contexts, including short-time Fourier transform computation
on neuromorphic hardware~\cite{Frady2022}, and for converting analog signals
into spike representations for downstream SNNs~\cite{Hille2022, Shaaban2024}. 
RF neurons have also been used as harmonic oscillators in feedforward SNNs for image, audio, and time-series classification, as well as related signal processing tasks~\cite{AlKhamissi2021, higuchi2024balanced, Frady2022, zhang2025dendritic}. Adaptation-based mechanisms in spiking networks offer a related, though distinct, route to temporally structured dynamics beyond standard LIF~\cite{baronig2025advancing}.
These works highlight the
potential of RF dynamics for efficient temporal representation and frequency
analysis. However, their integration into fixed-weight reservoir computing
frameworks remains largely unexplored. This work addresses this gap by
investigating oscillatory spiking neurons within the reservoir computing
paradigm.
\section{Methods}
\label{sec:model}

\subsection{Neuron Models}
\label{sec:neurons}

The LIF neuron is the standard building block of spiking reservoir
computing. Its membrane potential $v$ decays towards rest at a rate
controlled by the time constant $\tau_m$, while integrating incoming
input $I(t)$:
\begin{equation}
    \dot{v} = -\frac{v}{\tau_m} + I(t),
    \label{eq:lif_cont}
\end{equation}
where $\dot{v}$ denotes the time derivative of $v$. When $v$ crosses a threshold $\theta_{\text{\tiny LIF}}$, the neuron emits
a spike $s^{\text{\tiny LIF}} = 1$ and $v$ resets. The decay
term $-v/\tau_m$ acts as a leaky integrator: it smooths out rapid
fluctuations in the input, so the membrane potential reflects a
low-pass-filtered version of $I(t)$. As a consequence, the LIF
neuron responds most strongly to  low-frequency inputs, regardless of the
frequency content of the signal.

The Harmonic Resonate-and-Fire (HRF)
neuron~\cite{AlKhamissi2021,izhikevich2001resonate} replaces this LIF integration with the dynamics of a damped harmonic oscillator.
Its state is described by two coupled variables, a velocity $x$ and a
position $y$:
\begin{align}
    \dot{x} &= I(t) - \gamma \, y - \varepsilon \, x,
    \label{eq:hrf_x_cont} \\
    \dot{y} &= x.
    \label{eq:hrf_y_cont}
\end{align}
Here $\gamma > 0$ is the stiffness, which sets the natural oscillation
frequency $\omega = \sqrt{\gamma}$ of the neuron,
and $\varepsilon > 0$ is the damping, which controls how quickly
oscillations decay. The neuron emits a spike $s^{\text{\tiny HRF}} = 1$
when $y > \theta_{\text{\tiny RF}}$. The key difference from the LIF neuron
is that the restoring force $-\gamma y$ in Eq.~\eqref{eq:hrf_x_cont}
causes the membrane to oscillate rather than simply decay: the neuron
accumulates energy from inputs that arrive in phase with its natural
frequency $\omega$ and attenuates inputs at other frequencies. This gives
the HRF neuron band-pass rather than low-pass characteristics: it is
selectively sensitive to a specific frequency band determined by $\gamma$
and $\varepsilon$, rather than responding indiscriminately to all slow
input.

\subsection{The \NAME~Model}
\label{sec:arch}

\begin{figure}[t]
    \centering
    \includegraphics[width=0.65\textwidth]{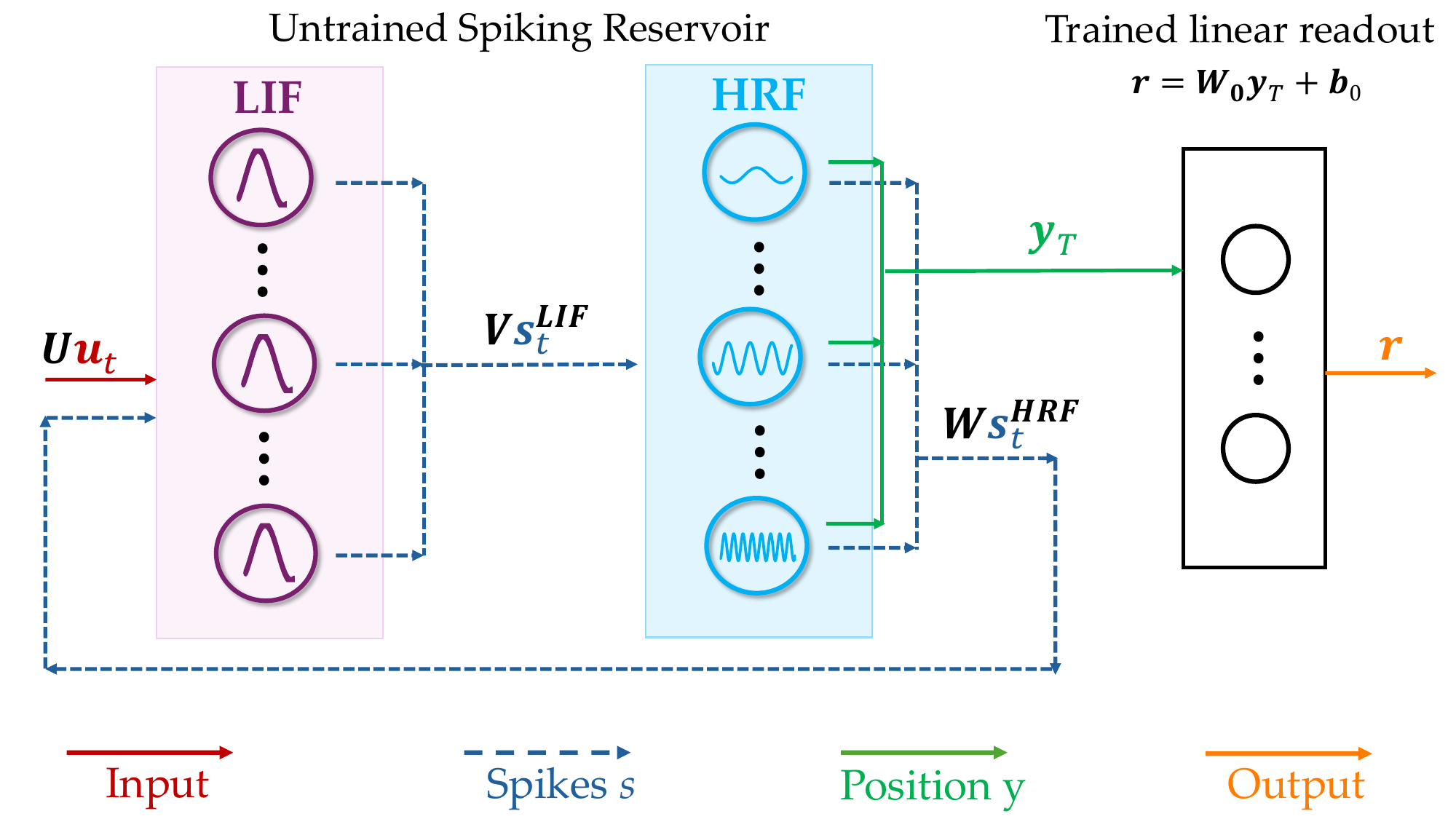}
\caption{%
    \textbf{Architecture of our proposed \NAME.}
    The external input $\mathbf{u}_t$ is projected into a LIF encoder population (left, pink) 
    via the fixed matrix $\mathbf{U}$. LIF spikes $\mathbf{s}_t^{\text{\tiny LIF}}$ are passed 
    to the HRF oscillatory reservoir (right, blue) via $\mathbf{V}$; each HRF neuron oscillates 
    at its own resonance frequency and feeds spikes $\mathbf{s}_t^{\text{\tiny HRF}}$ back to 
    the LIF layer via $\mathbf{W}$. All weights are fixed; only the linear readout 
    $\mathbf{r} = \mathbf{W}_o \mathbf{y}_T + \mathbf{b}_o$ is trained.
}
    \label{fig:architecture}
\end{figure}

Our proposed model, \NAME, organises $N$ neurons into two interacting populations, a
LIF encoder layer and an HRF oscillatory reservoir, all with fixed,
randomly initialised weights, as illustrated in 
Figure~\ref{fig:architecture}. Only a linear readout is optimised,
following the RC paradigm~\cite{jaeger2001echo,maass2002real}. Both
neuron models are simulated in discrete time via the Euler
method~\cite{atkinson1989introduction} with timestep $\Delta t$. The
complete discrete-time update at each timestep $t$ is

\begin{align}
    \mathbf{v}_{t+1} &= \mathbf{v}_t + \Delta t \left(
        -\frac{\mathbf{v}_t}{\tau_m}
        + \mathbf{U}\mathbf{u}_t
        + \mathbf{W}\mathbf{s}_t^{\text{\tiny HRF}}
        + \mathbf{b}
    \right),
    \label{eq:lif_full} \\[4pt]
    \mathbf{x}_{t+1} &= \mathbf{x}_t + \Delta t \left(
        \mathbf{V}\mathbf{s}_t^{\text{\tiny LIF}}
        - \boldsymbol{\gamma} \odot \mathbf{y}_t
        - \boldsymbol{\varepsilon} \odot \mathbf{x}_t
    \right),
    \label{eq:hrf_x_full} \\[4pt]
    \mathbf{y}_{t+1} &= \mathbf{y}_t + \Delta t \, \mathbf{x}_t,
    \label{eq:hrf_y_full}
\end{align}
where $\mathbf{u}_t \in \mathbb{R}^{N_\mathrm{inp}}$ is the external
input, $\mathbf{v}_t, \mathbf{x}_t, \mathbf{y}_t \in \mathbb{R}^N$ are
the LIF membrane potentials and the HRF velocity and position states
respectively. $\mathbf{s}_t^{\text{\tiny LIF}}, \mathbf{s}_t^{\text{\tiny
HRF}} \in \{0,1\}^N$ are the LIF and HRF spike vectors,
$\boldsymbol{\gamma}, \boldsymbol{\varepsilon} \in \mathbb{R}^N$ collect
the per-neuron stiffness and damping coefficients, and $\odot$ denotes
element-wise multiplication. $\mathbf{U} \in \mathbb{R}^{N \times
N_\mathrm{inp}}$ is the input matrix, $\mathbf{V} \in \mathbb{R}^{N
\times N}$ the LIF-to-HRF synaptic matrix, $\mathbf{W} \in
\mathbb{R}^{N \times N}$ the HRF-to-LIF feedback matrix, and
$\mathbf{b} \in \mathbb{R}^N$ a bias vector.

All weight matrices are randomly initialised and kept fixed throughout training. 
The input matrix $\mathbf{U}$ and bias $\mathbf{b}$ are scaled by a global 
input gain $\nu$, while the recurrent matrix $\mathbf{W}$ is rescaled to a 
target spectral radius $\rho$ following standard reservoir 
practice~\cite{jaeger2001echo,lukosevicius2012practical} (see 
Appendix~\ref{app:arch_details} for full initialisation details).
HRF neuron parameters $(\gamma_i, \varepsilon_i)$ are drawn independently 
across neurons, inducing a diversity of intrinsic time constants. This 
heterogeneity allows the reservoir to simultaneously represent temporal 
structure at multiple timescales, improving the expressivity of the fixed 
dynamics (see Appendix~\ref{app:arch_details}).
The timestep $\Delta t$ is treated as a hyperparameter rather than a fixed 
integration step~\cite{ceni2024ron}.
\NAME~is implemented in Python using standard machine learning libraries, 
and all hyperparameters are selected via random search on a held-out 
validation set (see Appendix~\ref{app:training_protocol} for the evaluation 
protocol and Appendix~\ref{app:hyperparams} for full parameter ranges and 
selected configurations).

\NAME~supports two families of readout strategies:
membrane-based readouts, which operate on the continuous HRF position
state $\mathbf{y}_t$, and spike-based readouts, which operate directly
on $\mathbf{s}_t^{\text{\tiny HRF}}$. This flexibility reflects the
practical deployment context: membrane-based readouts extract richer
temporal information from the oscillatory dynamics and are well-suited
to software or hybrid implementations, while spike-based readouts yield
a fully event-driven pipeline compatible with neuromorphic hardware
end-to-end. By default we use the final HRF state $\mathbf{y}_T$ as the readout 
feature, mapped to the output via $\mathbf{r} = \mathbf{W}_o \mathbf{y}_T 
+ \mathbf{b}_o$, where $\mathbf{W}_o \in \mathbb{R}^{N_\mathrm{out} \times N}$ 
and $\mathbf{b}_o \in \mathbb{R}^{N_\mathrm{out}}$ are the weights and biases 
of the output layer, optimised with regularisation strength $C$ treated as a 
hyperparameter. These are the \emph{only} trainable
parameters of \NAME. We conduct a systematic ablation over
membrane-based and spike-based readout strategies in
Section~\ref{sec:readout_ablation}, showing that the choice can be
adapted to the target deployment scenario without significant accuracy
loss.

\subsection{Energy Cost Model}
\label{sec:energy}

A central motivation for spiking reservoir computing is energy-efficient inference. To enable a consistent comparison between spiking and non-spiking models, we quantify computational cost using multiply-accumulate operations (MACs) for non-spiking networks and accumulate operations (ACs) for spiking networks~\cite{horowitz2014computing,yao2023spike}.
MACs correspond to dense multiplication-and-accumulation operations performed at every timestep over all connections, independent of neural activity. In contrast, ACs occur only when presynaptic spikes arrive and consist of adding synaptic contributions to the postsynaptic state, without runtime multiplication.
We measure spiking computation using synaptic operations (SOPs), defined as the number of spike-triggered synaptic transmissions, with each SOP corresponding directly to one AC event. This enables a unified comparison of spiking and non-spiking models under a common hardware energy model.
We convert MACs and ACs into energy using standard CMOS estimates ($E_{\mathrm{MAC}}$ and $E_{\mathrm{AC}}$). Following standard practice in the SNN literature, energy figures count synaptic operations, which scale as $\mathcal{O}(N^2)$ and dominate the cost of recurrent computation; the activity-independent $\mathcal{O}(N)$ neuron state updates, together with full details, are reported in Appendix~\ref{app:energy}.

\subsection{Reservoir Characterisation}
\label{sec:analysis}

A key hypothesis of this work is that oscillatory neuron dynamics in a spiking reservoir produce frequency-selective temporal representations that are more aligned with class structure than those of integrative dynamics, leading to improved classification. We test this through two complementary analyses on frozen reservoirs, using a controlled baseline that isolates the effect of neuron dynamics.

As a controlled baseline we introduce \LIFRC: a model identical
to \NAME~in every respect except that the HRF population (blue in Figure~\ref{fig:architecture}) is
replaced by LIF neurons with heterogeneous membrane time constants
$\tau_{m,i}$ and thresholds $\theta_{i}$, drawn per neuron from
log-uniform distributions. This mirrors the $\gamma_i / \varepsilon_i$
heterogeneity of the HRF reservoir, following prior work showing that
heterogeneous time constants improve
performance~\cite{deckers2022elsm,pereznieves2021neural}. An independent
hyperparameter search is conducted for \LIFRC~on each dataset using the
same protocol as for \NAME, ensuring the comparison reflects
the peak performance of each neuron model rather than an arbitrary
parameter transfer (see Appendix~\ref{app:lif_rc_details}). Any observed difference therefore has a single cause:
the oscillatory versus integrative nature of the reservoir dynamics.
The first analysis tests whether HRF neurons behave as band-pass filters in a full spiking reservoir, beyond the continuous-time, single-neuron regime of Eqs.~\eqref{eq:hrf_x_cont}--\eqref{eq:hrf_y_cont}. This property is not guaranteed to persist under discrete-time integration, recurrent coupling, and heterogeneous random connectivity. The second analysis evaluates whether such frequency selectivity yields representations that are intrinsically more discriminative than those of \LIFRC, even without readout training. This is a stricter criterion: frequency-selective responses do not necessarily translate into features that improve classification. Together, these analyses provide both an empirical characterisation of the HRF reservoir and evidence that this property is relevant for downstream tasks.

\paragraph{Frequency tuning and Q-factor.}
We measure the preferred input frequency  $f^*$ of each reservoir neuron by sweeping 30 log-spaced sinusoidal inputs through the full network and identifying the 
frequency maximising steady-state response power via the Discrete Fourier Transform (DFT). Filter sharpness is quantified by the Q-factor $Q = f^* / \Delta f_{-3\,\mathrm{dB}}$, where $\Delta f_{-3\,\mathrm{dB}}$ is the half-power bandwidth around the peak. A neuron is band-pass resonant ($Q > 0$) if both a lower and upper $-3$\,dB crossing exist, and integrative ($Q = 0$) otherwise. Being dimensionless and 
independent of $\Delta t$, $Q$ is directly comparable across models and 
datasets. Full sweep parameters are provided in Appendix~\ref{app:sweep}, where we also verify that applying a Hann window before the DFT, which reduces spectral leakage that could broaden the measured peaks, leaves all results unchanged.

\paragraph{Class discriminability without readout training (CKA).}
We measure how well reservoir states are
intrinsically aligned with the structure of the data using Centred
Kernel Alignment (CKA)~\cite{cristianini2002kernel,kornblith2019similarity},
without training any readout. Since CKA compares fixed-length representations, we summarise each trajectory by its temporal mean reservoir state: for each test sample $i$ we extract the
temporal mean reservoir state $\bar{\mathbf{y}}_i =
\frac{1}{T}\sum_{t=1}^{T} \mathbf{y}_t^{(i)} \in \mathbb{R}^N$, normalise each row to unit norm and stack these into $\mathbf{H} \in \mathbb{R}^{n 
\times N}$. Because the HRF position oscillates around an equilibrium, averaging partially cancels the very oscillatory component that distinguishes HRF from LIF dynamics, making this a conservative choice for \NAME. The feature kernel is $\mathbf{K}_H =
\mathbf{H}\mathbf{H}^\top$ and the label kernel is $(\mathbf{K}_Y)_{ij}
= \mathbf{1}[y_i = y_j]$. Both are centred with $\tilde{\mathbf{K}} =
\mathbf{H}_c \mathbf{K} \mathbf{H}_c$ where $\mathbf{H}_c =
\mathbf{I}_n - \frac{1}{n}\mathbf{1}\mathbf{1}^\top$, and CKA is
computed as
\begin{equation}
    \mathrm{CKA}(\mathbf{K}_H, \mathbf{K}_Y)
    = \frac{\langle \tilde{\mathbf{K}}_H,
            \tilde{\mathbf{K}}_Y \rangle_F}
           {\sqrt{\langle \tilde{\mathbf{K}}_H,
                  \tilde{\mathbf{K}}_H \rangle_F \,
                  \langle \tilde{\mathbf{K}}_Y,
                  \tilde{\mathbf{K}}_Y \rangle_F}},
    \label{eq:cka}
\end{equation}
where $\langle \cdot, \cdot \rangle_F$ is the Frobenius inner product.
A score of 1 indicates perfect 
class clustering; 0 indicates none. Further implementation
details, and controls showing that the results are robust to the number of samples $n$ and to discarding an initial transient of the trajectory, are provided in Appendix~\ref{app:cka_details}.

\section{Results}
\label{sec:results}

\subsection{HRF Reservoirs Function as Distributed Band-Pass Filters}
\label{sec:properties}

We apply the analyses of Section~\ref{sec:analysis} to compare frequency tuning and representational discriminability in \NAME~and \LIFRC~across four datasets chosen to span diverse input modalities and temporal regimes: sMNIST~\cite{lecun1998mnist} (sequential images, $T=784$, $N=800$), FordA~\cite{dau2019ucr} (univariate time series, $T=500$, $N=800$), SHD~\cite{cramer2020heidelberg} (audio spike trains, $T=250$, 700 channels, $N=3{,}000$), and DVS Gesture~\cite{amir2017dvs} (event-based vision, $T=200$, 2{,}048 channels, $N=3{,}000$).

\begin{figure}[t]
    \centering
    \includegraphics[width=0.80\textwidth]{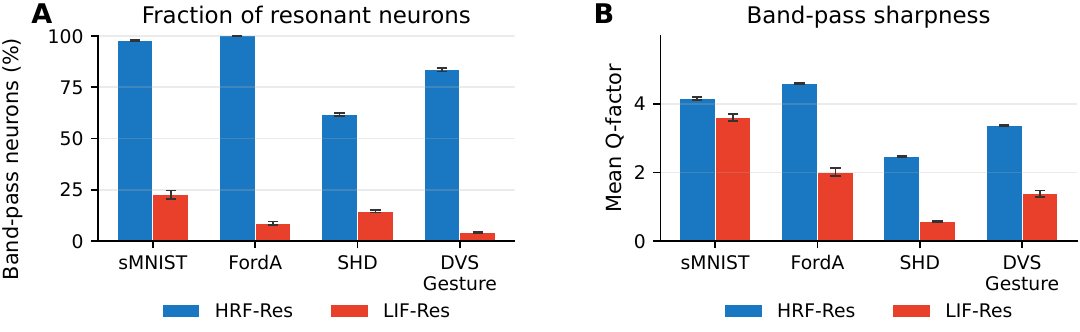}
    \caption{%
        \textbf{Band-pass properties of \NAME~vs.\ \LIFRC.}
        \textbf{(A)}~Fraction of resonant neurons ($Q > 0$) per dataset.
        \NAME~has 4--20$\times$ more resonant neurons than
        \LIFRC, confirming that band-pass dynamics survive in a full
        spiking reservoir setting.
        \textbf{(B)}~Mean Q-factor among resonant neurons. HRF neurons
        form sharper band-pass filters on all datasets, with the largest
        gaps on FordA ($4.6$ vs.\ $2.0$), DVS~Gesture ($3.4$ vs.\ $1.4$) and SHD ($2.5$ vs.\ $0.6$).
        Bars show mean~$\pm$~std over three seeds.
    }
    \label{fig:bandpass_summary}
\end{figure}

\begin{figure}[t]
    \centering
    \includegraphics[width=\textwidth]{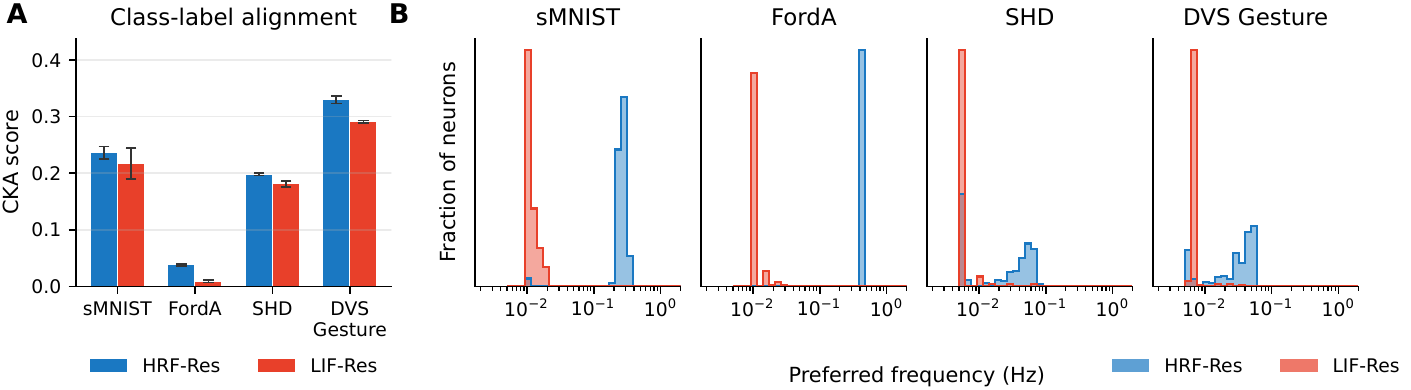}
    \caption{%
         \textbf{\NAME~produces more class-discriminative
        representations.}
        \textbf{(A)}~CKA between temporal mean reservoir states and
        category labels (mean~$\pm$~std over three seeds).
        \NAME~achieves higher CKA than \LIFRC~on all four datasets.
        \textbf{(B)}~Distribution of per-neuron preferred frequencies
        (log scale). HRF neurons show distributed tuning across a range of frequencies, consistent with a heterogeneous bank of band-pass filters;
        LIF neurons collapse to the lowest sweep frequency, consistent
        with pure integration.
    }
    \label{fig:repr_analysis_combined}
\end{figure}

\paragraph{Frequency tuning and Q-factor
(Figures~\ref{fig:bandpass_summary} and~\ref{fig:repr_analysis_combined}B).}
The difference between the two reservoirs is immediate. The fraction of
resonant neurons ($Q > 0$) is 98\%, 100\%, 62\%, and 83\% for \NAME~(sMNIST, FordA, SHD, DVS~Gesture), compared to 23\%, 9\%, 15\%, and 4\% for
\LIFRC~(mean over three seeds; standard deviations at most 2 percentage points). Beyond this expected gap, a key result is that many HRF neurons
retain sharp, well-defined resonance within the full recurrent network.
Among resonant units, Q-factors are consistently higher for HRF
(4.2 vs.\ 3.6, 4.6 vs.\ 2.0, 2.5 vs.\ 0.6, 3.4 vs.\ 1.4).
Preferred-frequency distributions further highlight this contrast:
HRF neurons span a broad range of tuned frequencies (standard deviation of $f^*$:
0.049, 0.041, 0.021, 0.016 Hz), whereas LIF neurons always collapse to the
lowest frequency (standard deviation: 0.002–0.008 Hz), consistent with pure integrative
behaviour. On FordA, HRF responses concentrate around 0.42 Hz, suggesting tuning to a dominant temporal scale in the signal.

\paragraph{Class discriminability
(Figure~\ref{fig:repr_analysis_combined}A).}
\NAME~achieves higher mean CKA than \LIFRC~on all four datasets. The gap is largest on FordA ($+322\%$), followed by DVS~Gesture ($+13\%$) and SHD ($+9\%$); on sMNIST ($+9\%$) the difference lies within the seed-to-seed variability. Absolute scores are low by design, as untrained reservoirs 
are not optimised for class discrimination. Notably, the ranking by CKA gap 
matches the ranking by absolute Q-factor gap (FordA $>$ DVS~Gesture $\approx$ SHD 
$>$ sMNIST), consistent with frequency selectivity driving representational 
discriminability.

Taken together, these analyses confirm that \NAME~functions as a
distributed bank of band-pass filters that project temporal inputs onto
frequency-structured, class-discriminative trajectories, a property that
\LIFRC~cannot produce despite being given equally heterogeneous
parameters and an independent hyperparameter search.
Section~\ref{sec:ablation} tests whether this representational advantage
translates to a classification improvement.

\subsection{HRF Neurons Improve Classification Across Modalities}
\label{sec:ablation}

We test whether the representational advantage of Section~\ref{sec:properties} 
translates to classification accuracy, using the same four datasets and reservoir sizes, and the 
final reservoir state at time $T$ as the readout feature. \NAME~outperforms \LIFRC~on three of four datasets, with
accuracy gains consistent with the predictions of
Section~\ref{sec:properties}~(Table~\ref{tab:lif_ablation}). The largest 
gap is on sMNIST ($+26.6\%$, $93.64\%$ vs.\ $67.00\%$), where HRF resonance 
is highest (98\% vs.\ 23\%). SHD follows with $+10.5$ ($83.48\%$ vs.\ 
$72.95\%$) and FordA with $+2.6$ ($69.72\%$ vs.\ $67.15\%$). DVS~Gesture is
the exception, where the two models perform comparably ($71.81 \pm
1.59\%$ vs.\ $73.48 \pm 1.12\%$) and the overlapping standard
deviations indicate no significant difference; we discuss this case in Section~\ref{sec:conclusion}. \NAME~is also more energy-efficient than \LIFRC~on most datasets despite achieving higher accuracy (energy estimated 
via the framework of Section~\ref{sec:energy}), as LIF neurons
appear to require higher firing rates to encode temporal information.
DVS~Gesture is again the exception.
These results confirm the hypothesis put forward in
Section~\ref{sec:properties}: equipping a spiking reservoir with
oscillatory HRF neurons produces measurably richer representations
than an equivalent LIF reservoir, and this representational advantage
translates directly to classification accuracy across diverse temporal
data modalities.

\begin{table}[t]
\centering
\caption{%
    \textbf{\NAME~vs.\ \LIFRC.} Both architectures are otherwise identical; 
    hyperparameters for \LIFRC~are independently re-tuned per dataset. 
    Results are mean~$\pm$~std over 5 seeds with final reservoir state readout. \NAME~outperforms or matches \LIFRC~on all four
    datasets, with accuracy gains up to $+26.6\%$ and lower energy
    consumption on most datasets. See Appendix~\ref{app:lif_rc_details} for \LIFRC~architecture
    details and Appendix~\ref{app:training_protocol} for the
    evaluation protocol.
}
\label{tab:lif_ablation}
\resizebox{0.63\columnwidth}{!}{%
\begin{tabular}{llcc}
\toprule
\textbf{Dataset} & \textbf{Model}
  & \textbf{Acc (\%)} & \textbf{Energy (J)} \\
\midrule
\multirow{2}{*}{sMNIST}
  & \textbf{\NAME} & $\mathbf{93.64 \pm 0.25}$
    & $(1.676 \pm 0.034) \times 10^{-4}$ \\
  & \LIFRC          & $67.00 \pm 1.04$
    & $(2.352 \pm 0.048) \times 10^{-4}$ \\
\midrule
\multirow{2}{*}{FordA}
  & \textbf{\NAME} & $\mathbf{69.72 \pm 0.68}$
    & $(0.790 \pm 0.015) \times 10^{-4}$ \\
  & \LIFRC          & $67.15 \pm 0.66$
    & $(1.478 \pm 0.038) \times 10^{-4}$ \\
\midrule
\multirow{2}{*}{SHD}
  & \textbf{\NAME} & $\mathbf{83.48 \pm 0.44}$
    & $(1.044 \pm 0.021) \times 10^{-3}$ \\
  & \LIFRC          & $72.95 \pm 0.53$
    & $(1.280 \pm 0.210) \times 10^{-3}$ \\
\midrule
\multirow{2}{*}{DVS Gesture}
  & \textbf{\NAME} & $71.81 \pm 1.59$
    & $(1.265 \pm 0.025) \times 10^{-3}$ \\
  & \LIFRC          & $\mathbf{73.48 \pm 1.12}$
    & $(0.953 \pm 0.074) \times 10^{-3}$ \\
\bottomrule
\end{tabular}
}
\end{table}

\subsection{Readout Strategy Ablation}
\label{sec:readout_ablation}

\begin{figure}[t]
    \centering
    \hspace*{-0.07\textwidth}
    \includegraphics[width=1.1\textwidth]{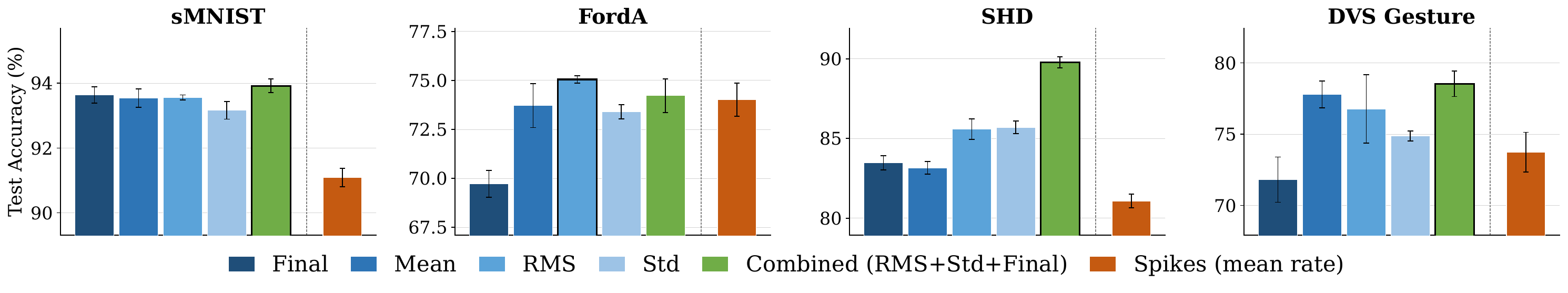}
    \caption{%
        \textbf{Readout strategy ablation.}
        Test accuracy (\%, mean $\pm$ std over 5 trials) for six
        readout strategies across four datasets. Blue bars: membrane-based
        readouts operating on the HRF position state $\mathbf{y}_t$,
        using the final state ($\mathbf{y}_T$), temporal mean, root mean
        square (RMS), or standard deviation (Std). Green bar: combined
        membrane-based readout concatenating RMS, Std, and final state
        ($3N$-dimensional). Orange bar: spike-based readout
        using the mean HRF spike rate. Bold outline marks the best accuracy per dataset.
        The best membrane-based readout outperforms the spike-based
        alternative on every dataset, and the combined readout achieves the best or
        near-best accuracy on all datasets.
    }
    \label{fig:readout_ablation}
\end{figure}

The choice of readout strategy is known to significantly affect
performance in reservoir computing~\cite{lukosevicius2012practical},
and carries practical implications for deployment: membrane-based readouts operating on $\mathbf{y}_t$ suit software 
or hybrid implementations, while spike-based readouts on $\mathbf{s}_t^{\text{\tiny 
HRF}}$ yield a fully event-driven neuromorphic pipeline. Even within membrane-based readouts, it is not clear which summary statistic 
of $\mathbf{y}_t$ best captures class-relevant information. We therefore 
compare six strategies (Figure~\ref{fig:readout_ablation}): five membrane-based 
operating on $\mathbf{y}_t$ (the final state $\mathbf{y}_T$, the temporal mean, 
the temporal root mean square, the temporal standard deviation, and a combined $3N$-dimensional 
vector concatenating 
the RMS, the standard deviation and the final state), and one spike-based readout 
using the mean firing rate of $\mathbf{s}_t^{\text{\tiny HRF}}$. Here $N$ is the number of 
reservoir neurons (Section~\ref{sec:arch}). The first four membrane-based modes 
and the spike-based mode all produce $N$-dimensional features, enabling a fair 
comparison.
On every dataset, the best membrane-based readout outperforms the spike-based alternative, indicating that the continuous oscillatory dynamics of $\mathbf{y}_t$ carry richer temporal information than mean spike rates. Among membrane-based 
modes, performance is stable across statistics on sMNIST and 
SHD, while temporal averaging outperforms the final state by $\sim$4\% on 
FordA, suggesting that aggregating over time reduces noise in univariate 
vibration signals. The combined readout achieves the best or near-best accuracy 
on all four datasets, at the cost of tripling the parameter count.

\subsection{Energy Efficiency of the Spiking Formulation}
\label{sec:energy_results}

Having established that oscillatory spiking dynamics outperform integrative 
ones, we ask whether the spiking formulation itself is justified relative to 
a non-spiking oscillatory reservoir, which is the relevant tradeoff when 
targeting deployment on neuromorphic hardware. We compare \NAME~against 
RON~\cite{ceni2024ron}, which shares the same core design but replaces LIF 
neurons with tanh-activated units and HRF neurons with continuous harmonic 
oscillators. For RON we use the hyperparameters 
reported in the original paper; for \NAME~we use the configurations selected 
via the random search described in Appendix~\ref{app:hyperparams}. Some accuracy cost is expected, as continuous activations carry 
more information per timestep than binary spikes; the question is whether the energy saving, quantified using the framework of 
Section~\ref{sec:energy}, is large enough to justify the spiking formulation. 
Both models are evaluated at matched reservoir size ($N = 800$) and readout 
(final reservoir state $\mathbf{y}_T$) on four of the benchmarks from the 
original RON evaluation~\cite{ceni2024ron}: sequential 
images (sMNIST, psMNIST, npCIFAR-10) and univariate time series (FordA).

As shown in Table~\ref{tab:ron_vs_sron}, \NAME~achieves 9--19$\times$ energy 
reduction over RON across all four benchmarks, with accuracy losses of only 
1.8--4.9 percentage points. The saving is structural: the dominant
$\mathcal{O}(N^2)$ recurrent cost is gated by spiking activity rather
than executed densely at every timestep, so it is expected to hold
broadly across tasks and deployment scenarios.
Including the activity-independent $\mathcal{O}(N)$ neuron state updates of both models, which add less than 10\% to the energy of \NAME, the reduction remains 9--17$\times$ (Appendix~\ref{app:energy}).
These results demonstrate that the spiking formulation of \NAME~achieves the energy efficiency required for edge deployment without
sacrificing the representational power of oscillatory reservoir dynamics,
offering a practical path to neuromorphic time series classification at
a fraction of the inference cost of non-spiking alternatives.

\definecolor{mygreen}{RGB}{220,240,220}
\definecolor{mygray}{RGB}{235,235,235}
\begin{table*}[t]
\centering
\caption{%
    \textbf{Energy efficiency: \NAME~vs.\ RON.} All experiments with $N = 800$ and final readout. 
    RON energy is computed from MAC counts (grey); \NAME~energy from SOPs 
    (green), which cost $\sim$5$\times$ less per operation than MACs in 45\,nm CMOS~\cite{horowitz2014computing}. \NAME~achieves 
    9--19$\times$ energy reduction over RON with accuracy losses of only 
    1.8--4.9 percentage points.
}
\label{tab:ron_vs_sron}
\noindent\makebox[\textwidth][c]{%
\resizebox{1.1\textwidth}{!}{%
\begin{tabular}{lcccccc c}
\toprule
& \multicolumn{3}{c}{\textbf{RON}~\cite{ceni2024ron}}
& \multicolumn{3}{c}{\textbf{\NAME~(ours)}}
& \\
\cmidrule(lr){2-4}\cmidrule(lr){5-7}
\textbf{Dataset}
  & \textbf{Acc (\%)}
  & \textbf{MACs ($\times10^8$)}
  & \cellcolor{mygray}\textbf{Energy ($\times10^{-4}$\,J)}
  & \textbf{Acc (\%)}
  & \textbf{SOPs ($\times10^8$)}
  & \cellcolor{mygreen}\textbf{Energy ($\times10^{-4}$\,J)}
  & \textbf{Reduction} \\
\midrule
sMNIST
  & $95.44\pm0.16$ & $5.024$ & \cellcolor{mygray}$23.11\pm0.69$
  & $93.64\pm0.25$ & $1.862\pm0.037$ & \cellcolor{mygreen}$1.676\pm0.042$
  & $13.8\times$ \\
psMNIST
  & $91.33\pm0.47$ & $4.921$ & \cellcolor{mygray}$22.64\pm0.68$
  & $87.96\pm0.33$ & $2.093\pm0.042$ & \cellcolor{mygreen}$1.884\pm0.046$
  & $12.0\times$ \\
FordA
  & $73.15\pm7.67$ & $3.204$ & \cellcolor{mygray}$14.74\pm0.44$
  & $69.72\pm0.68$ & $0.878\pm0.025$ & \cellcolor{mygreen}$0.790\pm0.015$
  & $18.7\times$ \\
npCIFAR-10
  & $38.50\pm0.32$ & $7.168$ & \cellcolor{mygray}$32.97\pm0.99$
  & $33.57\pm0.27$ & $3.830\pm1.200$ & \cellcolor{mygreen}$3.450\pm1.080$
  & $9.6\times$ \\
\bottomrule
\end{tabular}
}}
\end{table*}

\subsection{Comparison to Spiking Reservoir Baselines}
\label{sec:baselines}

We evaluate \NAME~against prior spiking reservoir methods on seven benchmarks 
chosen to match those reported by each baseline~\cite{deckers2022elsm, 
dey2022efficient, patino2022spinnaker, biswas2024icons}, enabling direct 
comparison: univariate time series (FordA, FordB, 
Wafer, Earthquakes~\cite{dau2019ucr}), audio spike trains 
(SHD~\cite{cramer2020heidelberg}), and neuromorphic vision 
(DVS~Gesture~\cite{amir2017dvs}, N-MNIST~\cite{orchard2015nmnist}). 
Reservoir sizes are matched to each baseline where a direct comparison exists; 
smaller reservoirs are used elsewhere to demonstrate efficiency. We use 
the combined readout for SHD, DVS~Gesture, N-MNIST, and larger-reservoir 
FordA; for FordB, Wafer, and Earthquakes we use the final-state readout, 
which achieves competitive results with fewer parameters.

\NAME~achieves competitive or state-of-the-art accuracy across all modalities 
(Table~\ref{tab:baseline_comparison}). On SHD, \NAME~with 6{,}000 neurons 
($89.78\%$) surpasses the 16{,}000-neuron ELSM~\cite{deckers2022elsm} at 
nearly $3\times$ fewer neurons. On FordB, Wafer, and Earthquakes, \NAME~with 
only 800 neurons outperforms all Dey et al.~\cite{dey2022efficient} baselines 
using up to $6\times$ more neurons. On FordA, \NAME~(large, $2{,}500$ neurons, 
$82.05\%$) outperforms the Dey et al.\ baseline at matched size, while even 
the small 800-neuron variant ($73.91\%$) remains competitive. On N-MNIST, 
\NAME~($97.13\%$) outperforms Pati\~{n}o-Saucedo et 
al.~\cite{patino2022spinnaker} ($94.43\%$) and approaches Biswas et 
al.~\cite{biswas2024icons} ($98.10\%$), who use a more complex ensemble 
architecture. On DVS~Gesture, \NAME~achieves $81.06\%$ versus $85.2\%$ for 
Biswas et al.\ at matched neuron count; the gap is attributable to their 
spatial and temporal ensemble architecture combining multiple reservoirs, 
rather than neuron count alone. All \NAME~results are averaged 
over 5 seeds with standard deviation, confirming consistent performance across 
a simple fixed random reservoir with no task-specific adaptations. Together, 
these results establish \NAME~as a competitive spiking reservoir model that 
generalises across diverse temporal modalities and input dimensionalities, 
whereas individual baselines typically cover only a subset of these settings.

\begin{table}[t]
\centering
\footnotesize
\caption{%
    \textbf{Comparison with spiking reservoir baselines.}
    Neurons: number of reservoir units. \NAME~results: mean~$\pm$~std 
    over 5 seeds. \NAME~achieves competitive or state-of-the-art accuracy 
    across all modalities, in most cases with fewer neurons 
    than prior methods.
}
\label{tab:baseline_comparison}
\resizebox{0.67\columnwidth}{!}{%
\begin{tabular}{llcc}
\toprule
\textbf{Dataset} & \textbf{Method} & \textbf{Neurons} & \textbf{Acc (\%)} \\
\midrule
\multirow{3}{*}{SHD}
  & \textbf{\NAME}                & 6{,}000  & $\mathbf{89.78\pm0.35}$ \\
  & Deckers et al.~\cite{deckers2022elsm}  & 16{,}000 & $89.30$ \\
  & Biswas et al.~\cite{biswas2024icons}   & 3{,}000  & $77.80$ \\
\midrule
\multirow{3}{*}{FordA}
  & \textbf{\NAME}              & 800     & $73.91\pm0.74$ \\
  & \textbf{\NAME}~(large)         & 2{,}500 & $\mathbf{82.05\pm0.55}$ \\
  & Dey et al.~\cite{dey2022efficient}  & 2{,}500 & $80.37$ \\
\midrule
\multirow{2}{*}{FordB}
  & \textbf{\NAME}              & 800     & $\mathbf{70.55\pm0.14}$ \\
  & Dey et al.~\cite{dey2022efficient}  & 2{,}500 & $64.32$ \\
\midrule
\multirow{2}{*}{Wafer}
  & \textbf{\NAME}              & 800     & $\mathbf{99.33\pm0.04}$ \\
  & Dey et al.~\cite{dey2022efficient}  & 5{,}000 & $98.85$ \\
\midrule
\multirow{2}{*}{Earthquakes}
  & \textbf{\NAME}              & 800     & $\mathbf{77.22\pm1.10}$ \\
  & Dey et al.~\cite{dey2022efficient}  & 4{,}500 & $71.94$ \\
\midrule
\multirow{3}{*}{N-MNIST}
  & \textbf{\NAME}                               & 3{,}600 & $97.13\pm0.12$ \\
  & Pati\~{n}o-Saucedo et al.~\cite{patino2022spinnaker}  & 4{,}096 & $94.43$ \\
  & Biswas et al.~\cite{biswas2024icons}                 & 3{,}600 & $\mathbf{98.10}$ \\
\midrule
\multirow{2}{*}{DVS Gesture}
  & \textbf{\NAME}               & 4{,}000 & $81.06\pm0.55$ \\
  & Biswas et al.~\cite{biswas2024icons}  & 4{,}000 & $\mathbf{85.20}$ \\
\bottomrule
\end{tabular}
}
\end{table}

\section{Conclusion}
\label{sec:conclusion}

We introduced \NAME, a spiking reservoir computing model based on Harmonic 
Resonate-and-Fire neurons, replacing conventional LIF dynamics with a 
heterogeneous population of oscillatory units. We showed that HRF neurons act 
as distributed band-pass filters within the reservoir, producing intrinsically 
more class-discriminative representations prior to readout training. This 
translates into consistent accuracy gains across diverse temporal benchmarks 
(up to $+26.6$\% over \LIFRC), while achieving up to $19\times$ lower 
inference energy than a non-spiking oscillatory baseline. Together, these 
results highlight frequency-selective dynamics as a powerful and underexplored 
computational principle for spiking reservoirs, enabling richer temporal 
representations without sacrificing efficiency.

Several limitations remain: energy estimates are model-based and not yet 
validated on neuromorphic hardware, the readout is trained offline, and a 
rigorous theoretical characterisation of HRF reservoir dynamics in terms of 
memory capacity, separation properties, and stability remains an open question. 
Additionally, the representational advantage of HRF neurons is expected to be 
most pronounced for signals with identifiable frequency structure; performance 
gains may be smaller for tasks where class-relevant information is not encoded 
in temporal frequency content. This is consistent with DVS~Gesture, where \NAME~does not improve over \LIFRC, and where class identity plausibly depends more on spatial trajectories than on stable temporal frequency content. Frequency selectivity also cuts both ways under distribution shift: \NAME~degrades more gracefully than \LIFRC~under additive noise, but is more sensitive to changes in the input sampling rate, which move the input spectrum away from the neurons' preferred bands (Appendix~\ref{app:robustness}). All results are based on classification benchmarks 
and generalisation to other settings remains to be explored. Two natural extensions are to learn the resonance parameters $(\gamma_i, \varepsilon_i)$, for instance with surrogate gradients, trading part of the training efficiency of reservoir computing for task-adapted frequency tuning, and to reintroduce the dynamic synapses of the original LSM formulation~\cite{maass2002real}, which we omitted, as is common in recent spiking reservoirs, to isolate the effect of neuron dynamics. Since short-term synaptic plasticity filters inputs at the synapse while HRF neurons filter them at the membrane, the two mechanisms may well be complementary.

\begin{ack}
This research received no specific external funding. The authors declare no competing interests.
\end{ack}

\newpage
\bibliographystyle{unsrt}

\bibliography{bibliography}

\appendix

\section{Experimental Details and Hyperparameter Search}
\label{app:hyperparams}
 
This appendix is self-contained: it provides all implementation
details, data preprocessing steps, training and evaluation procedures,
and hyperparameter configurations needed to reproduce the results in
the paper.
 
\subsection{Architecture Details}
\label{app:arch_details}
 
\paragraph{Weight initialisation.}
The input matrix $\mathbf{U} \in \mathbb{R}^{N \times N_{\mathrm{inp}}}$
is sampled entry-wise from $\mathcal{U}(0,1)$ and scaled by the input
scaling hyperparameter $\nu > 0$; the bias $\mathbf{b} \in \mathbb{R}^N$
is sampled from $\mathcal{U}(-1,1)$ and scaled by the same $\nu$.
The HRF-to-LIF feedback matrix $\mathbf{W} \in \mathbb{R}^{N \times N}$
is sampled from $\mathcal{U}(-2,2)$ and rescaled to a target spectral 
radius $\rho > 0$~\cite{jaeger2001echo,lukosevicius2012practical}.
The LIF-to-HRF synaptic matrix $\mathbf{V} \in \mathbb{R}^{N \times N}$
is sampled from $\mathcal{U}(-2,2)$ without spectral normalisation.
All matrices are fixed after initialisation and never updated.
\NAME~is implemented in PyTorch with a logistic 
regression readout from scikit-learn. Each input sequence is processed starting from a zero-initialised reservoir state.

\paragraph{Heterogeneous HRF parameters.}
Per-neuron stiffness and damping are drawn as
$\gamma_i \sim \mathcal{U}(\gamma - \Delta\gamma/2,\, \gamma + \Delta\gamma/2)$ and
$\varepsilon_i \sim \mathcal{U}(\varepsilon - \Delta\varepsilon/2,\, \varepsilon + \Delta\varepsilon/2)$,
with values clamped to be strictly positive.
The natural resonance frequency of neuron $i$ is $\omega_i = \sqrt{\gamma_i}$.
The centres $(\gamma, \varepsilon)$ and widths $(\Delta\gamma, \Delta\varepsilon)$ reported in Tables~\ref{tab:name_hparams_ablation}--\ref{tab:name_hparams_baseline} are treated as hyperparameters, inducing a diversity 
of intrinsic oscillation frequencies across the reservoir.
The timestep $\Delta t$ is treated as a hyperparameter rather than a fixed 
integration step, as we are interested in the computational properties of 
the discrete-time system in its own right~\cite{ceni2024ron}.

\paragraph{Spike generation.}
The position $y$ is not reset after a spike, so that the subthreshold oscillation continues undisturbed. To prevent a neuron from firing at every consecutive timestep while $y$ remains above threshold, the HRF threshold is made refractory: neuron $i$ spikes when $y_i > \theta_{\text{\tiny RF}} + r_i$, where $r_i$ is incremented by one at each spike and decays as $r_i \leftarrow r_i \, e^{-\Delta t/\tau_{\mathrm{ref}}}$ with $\tau_{\mathrm{ref}} = 0.25$.
 
\paragraph{Sparse connectivity.}
When the LIF$\to$HRF connectivity $p_V < 1$, a binary mask is sampled
i.i.d.\ with $\Pr(\text{entry} = 1) = p_V$ and applied entry-wise to
$\mathbf{V}$.
The HRF$\to$LIF connectivity is kept fully dense ($p_W = 1.0$) in all
experiments.
 
\paragraph{Sparse input projection.}
For high-dimensional inputs (SHD: 700 channels; DVS~Gesture: 2\,048
channels; N-MNIST: 578 channels), a sparse random projection is
applied to $\mathbf{U}$: a binary mask is sampled with each entry
set to 1 with probability $\rho_{\mathrm{inp}}$ (the input density
hyperparameter), and the masked weights are rescaled by
$1/\sqrt{\rho_{\mathrm{inp}}}$ to preserve the expected norm of the
input drive.
At least one active connection per neuron and per input channel is
ensured by rejection sampling.

\subsection{Energy Estimation Details}
\label{app:energy}

To quantify inference-time efficiency, we estimate computational cost using synaptic operations (SOPs)~\cite{horowitz2014computing,yao2023spike}. SOPs measure the number of spike-driven synaptic events and provide a hardware-relevant proxy for energy consumption in spiking systems.

On neuromorphic hardware, each presynaptic spike crossing a synapse corresponds to a single accumulate (AC) operation with energy cost
$E_{\mathrm{AC}} = 0.9\,\mathrm{pJ}$ (45\,nm CMOS). In contrast, non-spiking neural networks require multiply-accumulate (MAC) operations at every timestep, with cost
$E_{\mathrm{MAC}} = 4.6\,\mathrm{pJ}$~\cite{horowitz2014computing}.

For a two-population reservoir, the total number of synaptic operations per input sequence is given by:
\begin{equation}
    \mathrm{SOPs} =
    r_{\text{\tiny HRF}} \cdot N \cdot T \cdot C_{h \to l}
    \;+\;
    r_{\text{\tiny LIF}} \cdot N \cdot T \cdot C_{l \to h},
    \label{eq:sops_app}
\end{equation}
where $r_{\text{\tiny HRF}}$ and $r_{\text{\tiny LIF}}$ denote the mean firing rates of HRF and LIF populations, $N$ is the number of reservoir neurons, $T$ is the sequence length, and $C_{h \to l}$ and $C_{l \to h}$ denote the number of active synaptic connections in each direction.

This formulation reflects the event-driven nature of spiking computation: synaptic cost scales with spike activity rather than with full connectivity. In contrast, non-spiking models perform dense matrix multiplications at every timestep, leading to computation that scales with network size and sequence length regardless of input sparsity.

The internal HRF state dynamics require $\mathcal{O}(N)$ operations per timestep independent of spike activity, while recurrent synaptic communication is gated by sparse spiking events. As a result, the dominant recurrent cost scales with effective firing activity rather than full $N^2$ connectivity, which is the primary source of energy savings in spiking reservoir models.

All energy comparisons in the main paper are derived from SOP counts using the above formulation and the hardware constants defined above.

\paragraph{Neuron state updates.}
The SOP count above covers the event-driven synaptic cost. Neuron state updates are instead paid at every timestep, regardless of spiking activity. With the constant factors $(1-\Delta t\,\varepsilon_i)$, $\Delta t\,\gamma_i$ and $(1-\Delta t/\tau_{m,i})$ precomputed, each HRF neuron requires 3 MACs per timestep (Eqs.~\eqref{eq:hrf_x_full}--\eqref{eq:hrf_y_full}), as it updates two coupled state variables, while each LIF neuron requires 1 MAC (Eq.~\eqref{eq:lif_full}); bias additions and threshold comparisons are neglected. Per timestep, this amounts to $4N$ MACs for \NAME~(HRF reservoir and LIF encoder), $2N$ MACs for \LIFRC, and $3N$ MACs for RON, whose oscillators follow the same update (the $\tanh$ evaluation is not counted). These costs scale as $\mathcal{O}(N)$, whereas the synaptic cost scales as $\mathcal{O}(N^2)$.

Table~\ref{tab:energy_breakdown} reports the resulting estimates for the comparison of Table~\ref{tab:ron_vs_sron}. State updates add less than 10\% to the energy of \NAME~on all four datasets, and the reduction over RON remains 9--17$\times$. For the comparison of Table~\ref{tab:lif_ablation}, including state updates increases the energy of \NAME~by at most $0.12 \times 10^{-4}$\,J (sMNIST) and does not change the ranking between \NAME~and \LIFRC~on any dataset.

\begin{table}[h]
\centering
\caption{%
  \textbf{Energy including neuron state updates} ($\times 10^{-4}$\,J per sequence, $N=800$). ``Synaptic'' is the energy reported in Table~\ref{tab:ron_vs_sron}; ``State'' is the cost of the $\mathcal{O}(N)$ neuron state updates, computed as MACs.
}
\label{tab:energy_breakdown}
\small
\begin{tabular}{lccc ccc c}
\toprule
& \multicolumn{3}{c}{\textbf{RON}} & \multicolumn{3}{c}{\textbf{\NAME}} & \\
\cmidrule(lr){2-4}\cmidrule(lr){5-7}
\textbf{Dataset} & Synaptic & State & Total & Synaptic & State & Total & \textbf{Reduction} \\
\midrule
sMNIST     & 23.11 & 0.09 & 23.20 & 1.676 & 0.115 & 1.791 & $13.0\times$ \\
psMNIST    & 22.64 & 0.09 & 22.73 & 1.884 & 0.115 & 1.999 & $11.4\times$ \\
FordA      & 14.74 & 0.06 & 14.80 & 0.790 & 0.074 & 0.864 & $17.1\times$ \\
npCIFAR-10 & 32.97 & 0.11 & 33.08 & 3.450 & 0.147 & 3.597 & $9.2\times$ \\
\bottomrule
\end{tabular}
\end{table}

\subsection{Data Preprocessing}
\label{app:data}
 
\textbf{sMNIST / psMNIST.}
MNIST images ($28 \times 28$ pixels) are read pixel-by-pixel to form
sequences of length $T = 784$.
For psMNIST, a fixed random permutation (seed 0, independent of model
seed) is applied to the pixel order before each forward pass.
 
\textbf{npCIFAR-10.}
Each $32 \times 32$ RGB image is read row-by-row, producing $T = 32$
steps of 96-dimensional input ($32 \times 3$ channels).
Following~\cite{ceni2024ron}, 968 steps of fixed random Gaussian
noise (drawn once, shared across all trials) are appended, giving
$T = 1000$ total steps.
 
\textbf{FordA / FordB / Wafer / Earthquakes.}
Raw univariate sequences from the UCR archive~\cite{dau2019ucr} are
used without modification.
Sequence lengths are: FordA and FordB: $T = 500$; Wafer: $T = 152$;
Earthquakes: $T = 512$.
 
\textbf{SHD.}
Event streams are binned into $T = 250$ frames over a 1.4\,s window
using \texttt{tonic}, yielding sparse binary
tensors of shape $(250 \times 700)$.
 
\textbf{DVS Gesture.}
Events are binned into $T = 200$ frames; spatial downsampling by
factor 4 reduces $128 \times 128$ to $32 \times 32$ per polarity,
giving 2\,048 input channels per step.
 
\textbf{N-MNIST.}
Events are binned into $T = 30$ frames; spatial downsampling by
factor 2 reduces $34 \times 34$ to $17 \times 17$ per polarity,
giving 578 input channels.

\subsection{Training and Evaluation Protocol}
\label{app:training_protocol}
 
\paragraph{Readout.}
The only trained component is the linear readout: a logistic regression
classifier (scikit-learn) fitted with the L-BFGS
solver (max.\ 2\,000 iterations) and $\ell_2$ regularisation strength
$C$.
For SHD (20 classes) the multinomial objective is used; all other
datasets use one-vs-rest.
 
\paragraph{Data splits.}
For UCR datasets (FordA, FordB, Wafer, Earthquakes), the full
provided training partition is used for readout training.
For all other datasets the standard train/test splits are used.
Hyperparameter search uses a held-out validation split of 10\% of the
training set (stratified); the test set is never accessed during
search.
 
\paragraph{Evaluation.}
All results are mean\,$\pm$\,std over five independent trials.
Each trial uses a different random seed controlling all fixed weight
matrices ($\mathbf{U}$, $\mathbf{V}$, $\mathbf{W}$, $\mathbf{b}$)
and, for neuromorphic datasets, the sparse input projection mask.
The readout is re-fitted from scratch in each trial.
The same five seeds are used for \NAME~and \LIFRC~to ensure a controlled comparison.

\paragraph{Hyperparameter search.}
Each configuration is evaluated on the validation split with a single
seed; the best configuration is then re-evaluated with five seeds on
the full train/test split.
Parameters spanning multiple orders of magnitude ($\Delta t$, $\gamma$,
$\varepsilon$, $s_{\mathrm{inp}}$, $\theta_{\mathrm{enc}}$,
$\theta_{\mathrm{res}}$, $\rho_{\mathrm{inp}}$) are
sampled log-uniformly; all others uniformly.
The regularisation $C$ is sampled from a small discrete set.
The number of search trials per dataset is reported in the
corresponding table caption.
 
\paragraph{Compute.}
All experiments were run on a single NVIDIA A100 GPU (40\,GB).
Feature extraction per trial ranges from $\sim$10\,s (sMNIST,
$N = 800$) to $\sim$15\,min (N-MNIST, $N = 3600$).
Total compute across all experiments (search and final evaluation) is
approximately 500 GPU-hours.
 
\subsection{\LIFRC~Baseline}
\label{app:lif_rc_details}
 
\LIFRC~is architecturally identical to \NAME~except
that the HRF reservoir is replaced by a second LIF population.
It consists of a LIF encoder and a LIF reservoir, both of size $N$,
connected by the same weight matrices $\mathbf{U}$, $\mathbf{V}$,
$\mathbf{W}$ with the same initialisation and sparsity conventions as
\NAME.
The reservoir LIF neurons have heterogeneous membrane time constants
$\tau_{m,i}$ and thresholds $\theta_{\mathrm{res},i}$, drawn
per neuron from log-uniform distributions centred on hyperparameters
$\tau_m \pm \Delta\tau_m/2$ and
$\theta_{\mathrm{res}} \pm \Delta\theta_{\mathrm{res}}/2$
respectively, mirroring the $\gamma_i/\varepsilon_i$ heterogeneity of
\NAME.
Soft reset is used: upon spiking, the membrane potential is
decremented by the threshold rather than reset to zero.
The encoder time constant is fixed to $\tau_{\mathrm{enc}} = 20$
for all datasets, identical to \NAME.
An independent hyperparameter search is conducted for \LIFRC~on each
dataset using the same protocol as for \NAME, with
80 trials for SHD and 100 trials for all other datasets,
all using the final reservoir state for the readout.
Full search ranges and selected configurations are in
Table~\ref{tab:lif_rc_hparams}.
 
\begin{sidewaystable}[htbp]
\centering
\caption{%
  Hyperparameter search ranges and selected configurations for the
  \textbf{\LIFRC} baseline.
  All datasets use \texttt{readout\_mode\,=\,final} and the same logistic-regression readout as \NAME~(Appendix~\ref{app:training_protocol}).
  For each \emph{searched} parameter, the search range is shown in
  {\footnotesize\textit{italics}} above the selected value (bold).
  \emph{Fixed} parameters are reported without a range.
  Ranges are log-uniform for parameters spanning orders of magnitude,
  and uniform otherwise.
  The encoder membrane time constant is fixed to
  $\tau_{\mathrm{enc}} = 20$ for all datasets.
  The search comprised 80 random trials for SHD and 100 trials for
  all other datasets.
}
\label{tab:lif_rc_hparams}
\setlength{\tabcolsep}{4pt}
\renewcommand{\arraystretch}{1.55}
\scriptsize
\begin{tabular}{@{}l
    r
    r
    c
    c
    c
    c
    c
    c
    c
    c
    c
    c
    c
    c
    @{}}
\toprule
\multirow{2}{*}{\textbf{Dataset}}
  & \multicolumn{5}{c}{\textbf{Fixed parameters}}
  & \multicolumn{9}{c}{\textbf{Searched parameters}
      {\normalfont\footnotesize\ (range / \textbf{selected})}} \\
\cmidrule(lr){2-6}\cmidrule(lr){7-15}
  & {$N$}
  & {Steps}
  & {$\rho_{\mathrm{inp}}$}
  & {\makecell{LIF$\to$HRF\\conn.}}
  & {\makecell{HRF$\to$LIF\\conn.}}
  & {$\Delta t$}
  & {$\rho$}
  & {$s_{\mathrm{inp}}$}
  & {$\theta_{\mathrm{enc}}$}
  & {$\tau_m$}
  & {$\Delta\tau_m$}
  & {$\theta_{\mathrm{res}}$}
  & {$\Delta\theta_{\mathrm{res}}$}
  & {$C$} \\
\midrule

sMNIST
  & 800 & 784 & $1.0$ & $0.2$ & $1.0$
  & \makecell{{\footnotesize\textit{[0.03,\,0.15]}} \\ \textbf{0.034}}
  & \makecell{{\footnotesize\textit{[0.85,\,1.05]}} \\ \textbf{0.976}}
  & \makecell{{\footnotesize\textit{[0.5,\,5.0]}}   \\ \textbf{0.847}}
  & \makecell{{\footnotesize\textit{[0.01,\,0.20]}} \\ \textbf{0.151}}
  & \makecell{{\footnotesize\textit{[5,\,50]}}       \\ \textbf{36.19}}
  & \makecell{{\footnotesize\textit{[3,\,40]}}       \\ \textbf{3.60}}
  & \makecell{{\footnotesize\textit{[$10^{-3}$,\,0.05]}} \\ \textbf{0.00254}}
  & \makecell{{\footnotesize\textit{[$10^{-3}$,\,0.04]}} \\ \textbf{0.01180}}
  & \makecell{{\footnotesize\textit{\{$10^{-3}$,\,$10^{-2}$,\,$10^{-1}$,\,1,\,10\}}} \\ \textbf{0.01}}
  \\
\midrule

FordA
  & 800 & 500 & $1.0$ & $0.2$ & $1.0$
  & \makecell{{\footnotesize\textit{[0.02,\,0.30]}} \\ \textbf{0.068}}
  & \makecell{{\footnotesize\textit{[0.60,\,1.00]}} \\ \textbf{0.875}}
  & \makecell{{\footnotesize\textit{[0.3,\,5.0]}}   \\ \textbf{4.036}}
  & \makecell{{\footnotesize\textit{[0.01,\,0.30]}} \\ \textbf{0.146}}
  & \makecell{{\footnotesize\textit{[10,\,80]}}      \\ \textbf{36.68}}
  & \makecell{{\footnotesize\textit{[5,\,60]}}       \\ \textbf{25.85}}
  & \makecell{{\footnotesize\textit{[0.005,\,0.20]}} \\ \textbf{0.157}}
  & \makecell{{\footnotesize\textit{[0.002,\,0.15]}} \\ \textbf{0.01254}}
  & \makecell{{\footnotesize\textit{\{$10^{-3}$,\,$10^{-2}$,\,$10^{-1}$,\,1,\,10\}}} \\ \textbf{0.1}}
  \\
\midrule

SHD
  & 3000 & 250 & $0.036$ & $0.2$ & $1.0$
  & \makecell{{\footnotesize\textit{[0.05,\,1.0]}} \\ \textbf{0.084}}
  & \makecell{{\footnotesize\textit{[0.70,\,1.30]}} \\ \textbf{1.172}}
  & \makecell{{\footnotesize\textit{[0.02,\,1.0]}}  \\ \textbf{0.196}}
  & \makecell{{\footnotesize\textit{[0.03,\,0.50]}} \\ \textbf{0.419}}
  & \makecell{{\footnotesize\textit{[5,\,60]}}       \\ \textbf{13.01}}
  & \makecell{{\footnotesize\textit{[3,\,50]}}       \\ \textbf{2.38}}
  & \makecell{{\footnotesize\textit{[$10^{-3}$,\,0.10]}} \\ \textbf{0.09598}}
  & \makecell{{\footnotesize\textit{[$10^{-3}$,\,0.04]}} \\ \textbf{0.00375}}
  & \makecell{{\footnotesize\textit{\{$10^{-3}$,\,$10^{-2}$,\,$10^{-1}$,\,1\}}} \\ \textbf{0.01}}
  \\
\midrule

DVS Gesture
  & 3000 & 200 & $0.031$ & $1.0$ & $1.0$
  & \makecell{{\footnotesize\textit{[0.05,\,0.40]}} \\ \textbf{0.056}}
  & \makecell{{\footnotesize\textit{[0.85,\,1.30]}} \\ \textbf{0.866}}
  & \makecell{{\footnotesize\textit{[0.01,\,0.15]}} \\ \textbf{0.031}}
  & \makecell{{\footnotesize\textit{[0.10,\,3.0]}}  \\ \textbf{0.533}}
  & \makecell{{\footnotesize\textit{[2,\,100]}}      \\ \textbf{58.59}}
  & \makecell{{\footnotesize\textit{[1,\,80]}}       \\ \textbf{23.17}}
  & \makecell{{\footnotesize\textit{[$10^{-3}$,\,0.10]}} \\ \textbf{0.02224}}
  & \makecell{{\footnotesize\textit{[$10^{-3}$,\,0.08]}} \\ \textbf{0.00194}}
  & \makecell{{\footnotesize\textit{\{$10^{-4}$,\,3$\times10^{-4}$,\,$10^{-3}$,\,3$\times10^{-3}$,\,$10^{-2}$\}}} \\ \textbf{0.01}}
  \\
\bottomrule
\end{tabular}

\smallskip
\begin{minipage}{0.97\linewidth}
\footnotesize
\textit{Notation.}
$N$: reservoir size.
Steps: sequence length (timesteps).
$\rho_{\mathrm{inp}}$: input spike density (fraction of active input channels per step).
LIF$\to$HRF / HRF$\to$LIF conn.: connection density of $\mathbf{V}$ and $\mathbf{W}$ respectively (see text above).
$\Delta t$: simulation timestep.
$\rho$: spectral radius of $\mathbf{W}$.
$s_{\mathrm{inp}}$: input scaling factor $\nu$.
$\theta_{\mathrm{enc}}$: encoder LIF firing threshold.
$\tau_m$: centre of the log-uniform distribution of reservoir membrane time constants.
$\Delta\tau_m$: width of that distribution (heterogeneity).
$\theta_{\mathrm{res}}$: centre of the log-uniform distribution of reservoir firing thresholds.
$\Delta\theta_{\mathrm{res}}$: width of that distribution (heterogeneity).
$C$: inverse $\ell_2$ regularisation strength of the readout (selected from the discrete candidate set shown).

\end{minipage}

\end{sidewaystable}
 
\subsection{\NAME: Hyperparameter Configurations}
\label{app:name_search}

Tables~\ref{tab:name_hparams_ablation} and~\ref{tab:name_hparams_baseline} 
report full search ranges and selected configurations for \NAME. 
Table~\ref{tab:name_hparams_ablation} covers the six datasets used in the 
ablation studies (Sections~\ref{sec:ablation} and~\ref{sec:energy_results}): 
sMNIST, FordA, SHD, DVS~Gesture, psMNIST, and npCIFAR-10. 
Table~\ref{tab:name_hparams_baseline} covers the datasets used in the 
comparison against spiking reservoir baselines (Section~\ref{sec:baselines}): 
FordB, Wafer, Earthquakes, and N-MNIST, together with the larger-reservoir 
configurations of FordA, SHD, and DVS~Gesture.

For all datasets, hyperparameters were selected by random search on a 
held-out validation set (log-uniform sampling for parameters spanning orders 
of magnitude, uniform otherwise). The LIF$\to$HRF connectivity (fraction of 
non-zero entries in $\mathbf{V}$) and, for high-dimensional-input datasets, 
the input density ($\rho_{\mathrm{inp}}$) were included as searched 
parameters. The HRF$\to$LIF connectivity was kept fully dense ($1.0$) 
throughout, and the encoder membrane time constant fixed to 
$\tau_{\mathrm{enc}}=20$.

For datasets with two reservoir sizes, oscillator parameters ($\Delta t$, 
$\gamma$, $\varepsilon$, $\Delta\gamma$, $\Delta\varepsilon$, 
$\theta_{\mathrm{res}}$) are transferred from the smaller-$N$ search 
without modification (marked~$\dagger$); only size-sensitive parameters 
($s_{\mathrm{inp}}$, $\rho$, $\theta_{\mathrm{enc}}$, $p_V$, $C$) are 
re-searched. FordB, Wafer, and Earthquakes share the search space of FordA 
($N = 800$), differing only in $\theta_{\mathrm{enc}}$ (FordB, Wafer: 
1.0; Earthquakes: 0.2). The Combined readout ($3N$-dimensional concatenation 
of RMS, standard deviation, and final state of $\mathbf{y}_t$; 
Section~\ref{sec:readout_ablation}) is used for larger-reservoir 
configurations and N-MNIST; the Final readout for all ablation datasets. For each searched parameter 
we report the search range in square brackets (continuous range, sampled log-uniformly or uniformly as described in Appendix~\ref{app:training_protocol}) or curly braces 
(discrete set), followed by the selected value in bold.

\begin{sidewaystable}[htbp]
\centering
\caption{%
  \textbf{\NAME} hyperparameter search ranges and selected
  configurations for the ablation datasets.
  Each searched parameter is shown as
  {\footnotesize\textit{range}} / \textbf{selected}.
  Fixed parameters have no range.
}
\label{tab:name_hparams_ablation}
\setlength{\tabcolsep}{2pt}
\renewcommand{\arraystretch}{1.2}
\tiny
\begin{tabular}{@{}l
    r r c c c
    c c c c c c c c c c c
    @{}}
\toprule
\multirow{2}{*}{\textbf{Dataset}}
  & \multicolumn{5}{c}{\textbf{Fixed}}
  & \multicolumn{11}{c}{\textbf{Searched} {\normalfont\tiny (range / \textbf{selected})}} \\
\cmidrule(lr){2-6}\cmidrule(lr){7-17}
  & {$N$} & {Steps} & {$\rho_{\mathrm{inp}}$}
  & {\scriptsize\makecell{HRF$\to$LIF\\conn.}} & {Readout}
  & {\scriptsize\makecell{LIF$\to$HRF\\conn.}}
  & {$\Delta t$} & {$\gamma$} & {$\varepsilon$}
  & {$\Delta\gamma$} & {$\Delta\varepsilon$}
  & {$s_{\mathrm{inp}}$} & {$\rho$}
  & {$\theta_{\mathrm{enc}}$} & {$\theta_{\mathrm{res}}$}
  & {$C$} \\
\midrule

sMNIST & 800 & 784 & 1.0 & 1.0 & Final
  & \textit{\{.1,.2,.5,1\}} / \textbf{.2}
  & \textit{[.03,.15]} / \textbf{.042}
  & \textit{[1,5]} / \textbf{2.7}
  & \textit{[.02,.3]} / \textbf{.08}
  & \textit{[.5,3.5]} / \textbf{2.0}
  & \textit{[.1,1.5]} / \textbf{1.0}
  & \textit{[.5,5]} / \textbf{2.0}
  & \textit{[.85,1.05]} / \textbf{.99}
  & \textit{[.01,.2]} / \textbf{.05}
  & \textit{[$10^{-3}$,.05]} / \textbf{.005}
  & \textit{\{.001,.01,.1,1,10\}} / \textbf{1.0}
  \\
\midrule

FordA ($N$\!=\!800) & 800 & 500 & 1.0 & 1.0 & Final
  & \textit{\{.1,.2,.5,1\}} / \textbf{1.0}
  & \textit{[.005,.5]} / \textbf{.051}
  & \textit{[.01,10]} / \textbf{7.01}
  & \textit{[$10^{-3}$,1]} / \textbf{.153}
  & \textit{[.1,5]} / \textbf{3.01}
  & \textit{[0,.5]} / \textbf{.419}
  & \textit{[.01,10]} / \textbf{.625}
  & \textit{[.3,1.1]} / \textbf{.749}
  & \textit{[.005,1]} / \textbf{.082}
  & \textit{[$5\mathrm{e}{-4}$,.5]} / \textbf{.001}
  & \textit{\{.001,.01,.1,1,10\}} / \textbf{.1}
  \\
\midrule

SHD ($N$\!=\!3000) & 3000 & 250 & \textit{[.02,.3]}/\textbf{.036} & 1.0 & Final
  & \textit{\{.1,.2,.5,1\}} / \textbf{.2}
  & \textit{[.2,.8]} / \textbf{.223}
  & \textit{[.01,.5]} / \textbf{.036}
  & \textit{[.02,.12]} / \textbf{.060}
  & \textit{[.02,.5]} / \textbf{.268}
  & \textit{[0,.08]} / \textbf{.063}
  & \textit{[.02,1.8]} / \textbf{.230}
  & \textit{[.7,1.5]} / \textbf{1.16}
  & \textit{[.03,1]} / \textbf{1.00}
  & \textit{[$10^{-3}$,.02]} / \textbf{.013}
  & \textit{\{.001,.01,.1,1\}} / \textbf{.01}
  \\
\midrule

DVS ($N$\!=\!3000) & 3000 & 200 & \textit{[.01,.15]}/\textbf{.031} & 1.0 & Final
  & \textit{\{.1,.2,.5,1\}} / \textbf{1.0}
  & \textit{[.15,.4]} / \textbf{.259}
  & \textit{[.01,.15]} / \textbf{.046}
  & \textit{[.003,.08]} / \textbf{.035}
  & \textit{[.01,.15]} / \textbf{.130}
  & \textit{[.01,.10]} / \textbf{.099}
  & \textit{[.01,.12]} / \textbf{.113}
  & \textit{[1,1.6]} / \textbf{1.58}
  & \textit{[.3,3]} / \textbf{2.97}
  & \textit{[.005,.05]} / \textbf{.036}
  & \textit{\{$10^{-4}$,\ldots,.01\}} / \textbf{.01}
  \\
\midrule

psMNIST & 800 & 784 & 1.0 & 1.0 & Final
  & \textit{\{.1,.2,.5,1\}} / \textbf{.2}
  & \textit{[.03,.15]} / \textbf{.047}
  & \textit{[1,6]} / \textbf{2.62}
  & \textit{[.05,.8]} / \textbf{.240}
  & \textit{[1,5]} / \textbf{3.84}
  & \textit{[.5,3]} / \textbf{1.86}
  & \textit{[1,8]} / \textbf{3.67}
  & \textit{[1,2]} / \textbf{1.55}
  & \textit{[.01,.2]} / \textbf{.050}
  & \textit{[$10^{-3}$,.05]} / \textbf{.005}
  & \textit{\{.001,.01,.1,1,10\}} / \textbf{1.0}
  \\
\midrule

npCIFAR-10 & 800 & 1000 & 1.0 & 1.0 & Final
  & \textit{\{.1,.2,.5,1\}} / \textbf{.2}
  & \textit{[.01,.2]} / \textbf{.055}
  & \textit{[.01,2]} / \textbf{.115}
  & \textit{[.01,2]} / \textbf{.178}
  & \textit{[.1,2]} / \textbf{1.42}
  & \textit{[0,1]} / \textbf{.963}
  & \textit{[$10^{-3}$,.5]} / \textbf{.058}
  & \textit{[.5,1.5]} / \textbf{1.47}
  & \textit{[.01,1]} / \textbf{.124}
  & \textit{[$10^{-3}$,.5]} / \textbf{.154}
  & \textit{\{.001,.01,.1,1,10\}} / \textbf{1.0}
  \\

\bottomrule
\end{tabular}

\smallskip
\begin{minipage}{0.97\linewidth}
\tiny
\textit{Notation.}
$N$: reservoir size. Steps: sequence length.
$\rho_{\mathrm{inp}}$: input density (fraction of input channels per neuron; $1.0$\,=\,dense; shown as range/selected when searched).
LIF$\to$HRF / HRF$\to$LIF: connection density of $\mathbf{V}$ and $\mathbf{W}$.
$\Delta t$: timestep. $\gamma$/$\varepsilon$: centre of HRF stiffness/damping distributions ($\omega\!=\!\sqrt{\gamma}$).
$\Delta\gamma$/$\Delta\varepsilon$: widths (heterogeneity).
$s_{\mathrm{inp}}$: input scaling. $\rho$: spectral radius of $\mathbf{W}$.
$\theta_{\mathrm{enc}}$/$\theta_{\mathrm{res}}$: LIF encoder / HRF reservoir thresholds.
$C$: inverse $\ell_2$ regularisation strength of the readout.
\end{minipage}

\end{sidewaystable}

\begin{sidewaystable}[htbp]
\centering
\caption{%
  \textbf{\NAME} hyperparameter search ranges and selected
  configurations for the baseline comparison datasets
  (Section~\ref{sec:baselines}).
  Larger-reservoir variants of FordA, SHD, and DVS~Gesture reuse
  oscillator parameters from Table~\ref{tab:name_hparams_ablation}
  (marked~$\dagger$) and re-search only size-sensitive parameters.
  FordB, Wafer, and Earthquakes share the same search space as FordA
  ($N$\!=\!800).
  Each searched parameter is shown as
  {\footnotesize\textit{range}} / \textbf{selected}.
  Fixed parameters have no range.
  $^\dagger$Transferred from smaller-$N$ search.
  \emph{Combined}: $3N$-dimensional readout (RMS\,+\,Std\,+\,Final
  of $\mathbf{y}_t$; Section~\ref{sec:readout_ablation}).
}
\label{tab:name_hparams_baseline}
\setlength{\tabcolsep}{2pt}
\renewcommand{\arraystretch}{1.2}
\tiny
\begin{tabular}{@{}l
    r r c c c
    c c c c c c c c c c c
    @{}}
\toprule
\multirow{2}{*}{\textbf{Dataset}}
  & \multicolumn{5}{c}{\textbf{Fixed}}
  & \multicolumn{11}{c}{\textbf{Searched} {\normalfont\tiny (range / \textbf{selected})}} \\
\cmidrule(lr){2-6}\cmidrule(lr){7-17}
  & {$N$} & {Steps} & {$\rho_{\mathrm{inp}}$}
  & {\scriptsize\makecell{HRF$\to$LIF\\conn.}} & {Readout}
  & {\scriptsize\makecell{LIF$\to$HRF\\conn.}}
  & {$\Delta t$} & {$\gamma$} & {$\varepsilon$}
  & {$\Delta\gamma$} & {$\Delta\varepsilon$}
  & {$s_{\mathrm{inp}}$} & {$\rho$}
  & {$\theta_{\mathrm{enc}}$} & {$\theta_{\mathrm{res}}$}
  & {$C$} \\
\midrule

FordA ($N$\!=\!2500) & 2500 & 500 & 1.0 & 1.0 & Combined
  & \textit{[.001,.5]} / \textbf{.001}
  & $.051^\dagger$
  & $7.01^\dagger$
  & $.153^\dagger$
  & $3.01^\dagger$
  & $.419^\dagger$
  & \textit{[.2,4]} / \textbf{2.13}
  & \textit{[.6,1.05]} / \textbf{.985}
  & \textit{[.5,2.5]} / \textbf{1.21}
  & $.001^\dagger$
  & \textit{[$5\mathrm{e}{-4}$,.05]} / \textbf{.0005}
  \\
\midrule

SHD ($N$\!=\!6000) & 6000 & 250 & $.036^\dagger$ & 1.0 & Combined
  & $1.0^\dagger$
  & $.223^\dagger$
  & $.036^\dagger$
  & $.060^\dagger$
  & $.268^\dagger$
  & $.063^\dagger$
  & $.230^\dagger$
  & $1.16^\dagger$
  & $1.00^\dagger$
  & $.013^\dagger$
  & $.01^\dagger$
  \\
\midrule

DVS ($N$\!=\!4000) & 4000 & 200 & \textit{[.01,.10]}/\textbf{.011} & 1.0 & Combined
  & \textit{\{.1,.2,.5,1\}} / \textbf{.2}
  & $.259^\dagger$
  & $.046^\dagger$
  & $.035^\dagger$
  & $.130^\dagger$
  & $.099^\dagger$
  & \textit{[.02,.3]} / \textbf{.112}
  & \textit{[.85,1.6]} / \textbf{1.38}
  & \textit{[.5,5]} / \textbf{4.44}
  & \textit{[.02,.2]} / \textbf{.093}
  & \textit{\{$10^{-4}$,3e-4,$10^{-3}$,3e-3,.01,.03\}} / \textbf{.01}
  \\
\midrule

FordB & 800 & 500 & 1.0 & 1.0 & Final
  & \textit{\{.1,.2,.5,1\}} / \textbf{.2}
  & \textit{[.005,.5]} / \textbf{.2}
  & \textit{[.01,10]} / \textbf{1.88}
  & \textit{[$10^{-3}$,1]} / \textbf{.022}
  & \textit{[.1,5]} / \textbf{2.64}
  & \textit{[0,.5]} / \textbf{.068}
  & \textit{[.01,10]} / \textbf{1.76}
  & \textit{[.3,1.1]} / \textbf{.95}
  & \textit{[.005,1]} / \textbf{1.0}
  & \textit{[$5\mathrm{e}{-4}$,.5]} / \textbf{.005}
  & \textit{\{.001,.01,.1,1,10\}} / \textbf{1.0}
  \\
\midrule

Wafer & 800 & 152 & 1.0 & 1.0 & Final
  & \textit{\{.1,.2,.5,1\}} / \textbf{.2}
  & \textit{[.005,.5]} / \textbf{.2}
  & \textit{[.01,10]} / \textbf{1.88}
  & \textit{[$10^{-3}$,1]} / \textbf{.022}
  & \textit{[.1,5]} / \textbf{2.64}
  & \textit{[0,.5]} / \textbf{.068}
  & \textit{[.01,10]} / \textbf{1.76}
  & \textit{[.3,1.1]} / \textbf{.95}
  & \textit{[.005,1]} / \textbf{1.0}
  & \textit{[$5\mathrm{e}{-4}$,.5]} / \textbf{.005}
  & \textit{\{.001,.01,.1,1,10\}} / \textbf{1.0}
  \\
\midrule

Earthquakes & 800 & 512 & 1.0 & 1.0 & Final
  & \textit{\{.1,.2,.5,1\}} / \textbf{.2}
  & \textit{[.005,.5]} / \textbf{.2}
  & \textit{[.01,10]} / \textbf{1.88}
  & \textit{[$10^{-3}$,1]} / \textbf{.022}
  & \textit{[.1,5]} / \textbf{2.64}
  & \textit{[0,.5]} / \textbf{.068}
  & \textit{[.01,10]} / \textbf{1.76}
  & \textit{[.3,1.1]} / \textbf{.95}
  & \textit{[.005,1]} / \textbf{.2}
  & \textit{[$5\mathrm{e}{-4}$,.5]} / \textbf{.005}
  & \textit{\{.001,.01,.1,1,10\}} / \textbf{1.0}
  \\
\midrule

N-MNIST & 3600 & 30 & \textit{[.01,.20]}/\textbf{.018} & 1.0 & Combined
  & \textit{\{.1,.2,.5,1\}} / \textbf{.2}
  & .109
  & .109
  & .021
  & .362
  & .083
  & \textit{[.02,.5]} / \textbf{.210}
  & \textit{[.8,1.6]} / \textbf{1.05}
  & \textit{[.05,.5]} / \textbf{.494}
  & \textit{[.01,.15]} / \textbf{.058}
  & \textit{\{.001,.005,.01,.05,.1,.5\}} / \textbf{.05}
  \\

\bottomrule
\end{tabular}

\smallskip
\begin{minipage}{0.97\linewidth}
\tiny
\textit{Notation} as in Table~\ref{tab:name_hparams_ablation}.
For SHD ($N$\!=\!6000), all parameters are transferred from the $N$\!=\!3000 search; only reservoir size and readout mode changed.
FordB, Wafer, and Earthquakes share the same search space as FordA ($N$\!=\!800); $\theta_{\mathrm{enc}}$ differs: $1.0$ for FordB and Wafer, $0.2$ for Earthquakes.
For DVS ($N$\!=\!4000) and N-MNIST, $\rho_{\mathrm{inp}}$ was also searched (shown in the fixed column as range/selected).
N-MNIST uses spatial downsampling factor\,2 (17$\times$17$\times$2\,=\,578 input channels).
\end{minipage}

\end{sidewaystable}

\newpage

\section{Reservoir Characterisation: Supplementary Details}
\label{app:sweep}

\subsection{Frequency Sweep Parameters}
\label{app:sweep_params}

The frequency sweep analysis sweeps 30 log-spaced frequencies per dataset.
For each frequency $f$, a sinusoidal input $x(t) = \sin(2\pi f \cdot t
\cdot \Delta t)$ is broadcast identically across all $N_\mathrm{inp}$
input channels and the network is run for $T_\mathrm{sweep}$ timesteps.
The first half of the trajectory is discarded as transient; each neuron's
preferred frequency $f^*$ is identified as the frequency maximising its
steady-state response power, estimated via the Discrete Fourier Transform
(DFT) of the second half. The same sweep range $[f_{\min}, f_{\max}]$ is
applied to both \NAME~and \LIFRC~for each dataset; each model
uses its own independently tuned $\Delta t$.

\begin{table}[h]
\centering
\caption{Frequency sweep parameters and model $\Delta t$ values per dataset.}
\label{tab:sweep_params}
\begin{tabular}{lcccccc}
\toprule
Dataset & $f_{\min}$ (Hz) & $f_{\max}$ (Hz) & $T_\mathrm{sweep}$
        & $\Delta t$ [\NAME] & $\Delta t$ [\LIFRC] \\
\midrule
sMNIST      & 0.01  & 2.0 & 6{,}000  & 0.042 & 0.034 \\
FordA       & 0.01  & 2.0 & 6{,}000  & 0.051 & 0.068 \\
SHD         & 0.001 & 2.0 & 50{,}000 & 0.223 & 0.084 \\
DVS Gesture & 0.001 & 2.0 & 50{,}000 & 0.259 & 0.056 \\
\bottomrule
\end{tabular}
\end{table}

The larger $T_\mathrm{sweep}$ and smaller $f_{\min}$ for SHD and DVS
Gesture reflect the slower intrinsic timescales of these datasets: the
hyperparameter search selects larger $\Delta t$ values to match the
longer temporal dependencies present in audio spike trains and
event-camera streams. Using a shorter sweep or higher $f_{\min}$ for
these datasets would fail to resolve the low-frequency resonances of the
HRF neurons, underestimating their Q-factors.

The Q-factor is computed per neuron from the response power matrix
$R \in \mathbb{R}^{N_\mathrm{freqs} \times N}$ as follows. For each
neuron $j$, the peak response $P_\mathrm{peak}$ is identified at its
preferred frequency $f^*$. The half-power bandwidth $\Delta
f_{-3\,\mathrm{dB}}$ is estimated by finding the lower and upper
frequency crossings where $R_{fj} = P_\mathrm{peak}/2$, using linear
interpolation on the log-frequency axis. $Q = f^* / \Delta
f_{-3\,\mathrm{dB}}$ is set to zero if either crossing does not exist,
indicating purely integrative (low-pass) behaviour with no well-defined
resonance peak. We emphasise that the sweep provides a mechanistic
characterisation of each reservoir's dynamical repertoire and does not
constitute a direct model-to-model comparison of absolute frequency
scales, since the two models have independently tuned $\Delta t$ values.

\paragraph{Effect of spectral windowing.}
The results in Section~\ref{sec:properties} compute the DFT on the steady-state trajectory without an explicit window. To verify that spectral leakage does not affect our conclusions, we repeated the analysis applying a Hann window before the DFT (Table~\ref{tab:hann}). Windowing shifts the estimates slightly but leaves every qualitative conclusion unchanged: on all four datasets, \NAME~has a far larger fraction of resonant neurons, higher Q-factors, and a broader distribution of preferred frequencies than \LIFRC.

\begin{table}[h]
\centering
\caption{Frequency-selectivity analysis with a Hann window applied before the DFT (mean~$\pm$~std over three seeds).}
\label{tab:hann}
\small
\begin{tabular}{llccc}
\toprule
\textbf{Dataset} & \textbf{Model} & \textbf{Resonant (\%)} & \textbf{Mean $Q$} & \textbf{Std of $f^*$ (Hz)} \\
\midrule
\multirow{2}{*}{sMNIST}      & \NAME  & $89.0 \pm 1.1$  & $4.38 \pm 0.04$ & $0.0829 \pm 0.0029$ \\
                             & \LIFRC & $16.7 \pm 2.8$  & $3.67 \pm 0.14$ & $0.0018 \pm 0.0001$ \\
\midrule
\multirow{2}{*}{FordA}       & \NAME  & $100.0 \pm 0.1$ & $4.62 \pm 0.03$ & $0.0395 \pm 0.0012$ \\
                             & \LIFRC & $3.8 \pm 0.6$   & $1.02 \pm 0.03$ & $0.0030 \pm 0.0005$ \\
\midrule
\multirow{2}{*}{SHD}         & \NAME  & $60.5 \pm 0.7$  & $2.49 \pm 0.02$ & $0.0238 \pm 0.0001$ \\
                             & \LIFRC & $5.7 \pm 0.1$   & $0.63 \pm 0.01$ & $0.0075 \pm 0.0005$ \\
\midrule
\multirow{2}{*}{DVS Gesture} & \NAME  & $80.3 \pm 1.0$  & $3.41 \pm 0.01$ & $0.0157 \pm 0.0001$ \\
                             & \LIFRC & $1.6 \pm 0.2$   & $1.49 \pm 0.07$ & $0.0036 \pm 0.0004$ \\
\bottomrule
\end{tabular}
\end{table}

\subsection{CKA: Implementation Details}
\label{app:cka_details}

We use the linear kernel throughout, following~\cite{kornblith2019similarity}.
For each test sample $i$, the temporal mean reservoir state
$\bar{\mathbf{y}}_i = \frac{1}{T}\sum_{t=1}^{T} \mathbf{y}_t^{(i)} \in
\mathbb{R}^N$ is computed over the full sequence, starting from a zero-initialised reservoir state. Each row is then normalised to unit norm before kernel
computation, removing differences in overall activation magnitude between
the two models and ensuring the comparison reflects representational
geometry rather than scale. The feature kernel is $\mathbf{K}_H =
\mathbf{H}\mathbf{H}^\top$ where $\mathbf{H} \in \mathbb{R}^{n \times N}$
stacks the normalised mean states, and the label kernel is
$(\mathbf{K}_Y)_{ij} = \mathbf{1}[y_i = y_j]$. Both kernels are centred
with the centering matrix $\mathbf{H}_c = \mathbf{I}_n -
\frac{1}{n}\mathbf{1}\mathbf{1}^\top$ and CKA is computed as in
Eq.~\eqref{eq:cka}. If either $\langle \tilde{\mathbf{K}}_H,
\tilde{\mathbf{K}}_H \rangle_F$ or $\langle \tilde{\mathbf{K}}_Y,
\tilde{\mathbf{K}}_Y \rangle_F$ is smaller than $10^{-10}$, CKA is set
to zero to avoid numerical instability.

We use $n = 500$ randomly sampled test examples for all datasets except
DVS~Gesture, for which we use the full test set ($n = 264$). The choice
of $n = 500$ is a practical constraint: computing the full $n \times n$
kernel matrix becomes memory-intensive for larger $n$. Both models are
evaluated at their respective best hyperparameter configurations as
determined by the random search described in
Appendix~\ref{app:hyperparams}, ensuring the comparison reflects the
best achievable representational quality for each neuron model.

\paragraph{Choice of summary statistic.}
CKA~\cite{kornblith2019similarity} compares fixed-length representations, so applying it to reservoir trajectories requires a summary statistic. We use the temporal mean as a conservative choice: since the HRF position $\mathbf{y}_t$ oscillates around an equilibrium, averaging partially cancels the oscillatory component that distinguishes HRF from LIF dynamics. Statistics that retain more of this component, such as the RMS or the combined statistic of Section~\ref{sec:readout_ablation}, would be expected to accentuate rather than diminish the differences reported in Section~\ref{sec:properties}.

\paragraph{Sensitivity to sample size.}
Linear CKA estimates can be noisy for small $n$. For a fixed reservoir per model, we recomputed CKA over repeated random subsamples of the test set of increasing size, up to the full test set (Table~\ref{tab:cka_n}). The variability decreases markedly with $n$, the estimate at our reported sample sizes is close to the full-test-set value, and the \NAME--\LIFRC~gap is positive and of comparable magnitude at every sample size, including the smallest.

\begin{table}[h]
\centering
\caption{CKA as a function of the number of test samples $n$ (mean~$\pm$~std over repeated random subsamples; one fixed reservoir per model). ``Full'': entire test set (1{,}320 / 10{,}000 / 2{,}264 / 264 samples for FordA / sMNIST / SHD / DVS~Gesture), for which only one sample is possible. DVS~Gesture has fewer than 500 test samples.}
\label{tab:cka_n}
\resizebox{\columnwidth}{!}{%
\begin{tabular}{lcccccccc}
\toprule
& \multicolumn{2}{c}{\textbf{FordA}} & \multicolumn{2}{c}{\textbf{sMNIST}} & \multicolumn{2}{c}{\textbf{SHD}} & \multicolumn{2}{c}{\textbf{DVS Gesture}} \\
\cmidrule(lr){2-3}\cmidrule(lr){4-5}\cmidrule(lr){6-7}\cmidrule(lr){8-9}
$n$ & \NAME & \LIFRC & \NAME & \LIFRC & \NAME & \LIFRC & \NAME & \LIFRC \\
\midrule
100  & $.0555\pm.0217$ & $.0211\pm.0104$ & $.2628\pm.0244$ & $.2238\pm.0267$ & $.2632\pm.0236$ & $.2380\pm.0359$ & $.3552\pm.0333$ & $.3245\pm.0286$ \\
200  & $.0357\pm.0115$ & $.0101\pm.0044$ & $.2298\pm.0244$ & $.1922\pm.0264$ & $.2294\pm.0149$ & $.2049\pm.0174$ & $.3253\pm.0125$ & $.2989\pm.0120$ \\
500  & $.0274\pm.0039$ & $.0053\pm.0025$ & $.2217\pm.0101$ & $.1781\pm.0135$ & $.2063\pm.0084$ & $.1838\pm.0137$ & -- & -- \\
1000 & $.0268\pm.0035$ & $.0034\pm.0009$ & $.2166\pm.0081$ & $.1761\pm.0102$ & $.2018\pm.0072$ & $.1804\pm.0069$ & -- & -- \\
Full & $.0263$ & $.0029$ & $.2111$ & $.1729$ & $.1959$ & $.1732$ & $.3203$ & $.2914$ \\
\bottomrule
\end{tabular}
}
\end{table}

\paragraph{Sensitivity to the initial transient.}
Since the temporal mean includes the initial transient from the zero-initialised state, we recomputed CKA after discarding the first 10\%, 25\%, and 50\% of each trajectory (Table~\ref{tab:cka_washout}). CKA changes only marginally with the washout length, and \NAME~retains a higher score than \LIFRC~in every setting. The classification experiments with the final-state readout are unaffected by washout by construction.

\begin{table}[h]
\centering
\caption{CKA after discarding an initial fraction of each trajectory (washout; one fixed reservoir per model). Washout steps at 10/25/50\%: FordA 50/125/250; sMNIST 78/196/392; SHD 25/62/125; DVS~Gesture 20/50/100.}
\label{tab:cka_washout}
\small
\begin{tabular}{lcccccccc}
\toprule
& \multicolumn{2}{c}{\textbf{FordA}} & \multicolumn{2}{c}{\textbf{sMNIST}} & \multicolumn{2}{c}{\textbf{SHD}} & \multicolumn{2}{c}{\textbf{DVS Gesture}} \\
\cmidrule(lr){2-3}\cmidrule(lr){4-5}\cmidrule(lr){6-7}\cmidrule(lr){8-9}
Washout & \NAME & \LIFRC & \NAME & \LIFRC & \NAME & \LIFRC & \NAME & \LIFRC \\
\midrule
0\%  & .0399 & .0078 & .2708 & .2144 & .2143 & .1836 & .3509 & .2835 \\
10\% & .0414 & .0084 & .2707 & .2144 & .2145 & .1829 & .3509 & .2832 \\
25\% & .0365 & .0086 & .2723 & .2155 & .2143 & .1832 & .3509 & .2825 \\
50\% & .0301 & .0069 & .2720 & .2200 & .2142 & .1881 & .3507 & .2819 \\
\bottomrule
\end{tabular}
\end{table}

\section{Robustness to Temporal Perturbations}
\label{app:robustness}

The band-pass mechanism of Section~\ref{sec:properties} makes two opposite predictions under distribution shift: frequency selectivity should suppress broadband perturbations, but should make the reservoir more sensitive to shifts of the input spectrum. We test both on FordA and sMNIST, keeping trained reservoirs and readouts fixed and perturbing inputs only at test time (three independently initialised reservoirs per model). We consider additive zero-mean Gaussian noise with standard deviation $\sigma$, and sampling-rate shifts implemented by linear interpolation, where a rate below (above) $1.0$ slows down (speeds up) the input sequence. Time warping is not considered, as it lacks a single standard operationalisation.

Under additive noise (Table~\ref{tab:noise}), \NAME~degrades more gracefully than \LIFRC: at $\sigma = 0.5$, \NAME~loses 5.6 percentage points on FordA versus 14.4 for \LIFRC, and on sMNIST \NAME~retains above-chance accuracy at noise levels at which \LIFRC~drops to chance. Under sampling-rate shifts (Table~\ref{tab:rate}) the trend reverses: on sMNIST, \NAME~drops to near-chance accuracy for every shift tested, whereas \LIFRC~degrades less, and on FordA the two models degrade similarly. Both observations follow from the same mechanism: a bank of band-pass filters rejects energy outside its preferred bands, but a rate change moves the informative content of the signal away from those bands. Neither model, however, is robust to sampling-rate mismatch, and both would require retraining or adaptation in that setting.

\begin{table}[h]
\centering
\caption{Test accuracy (\%) under additive Gaussian input noise of standard deviation $\sigma$ (mean~$\pm$~std over three reservoirs). The $\sigma = 0$ row reproduces the clean results of Table~\ref{tab:lif_ablation} (five seeds).}
\label{tab:noise}
\small
\begin{tabular}{lcccc}
\toprule
& \multicolumn{2}{c}{\textbf{FordA}} & \multicolumn{2}{c}{\textbf{sMNIST}} \\
\cmidrule(lr){2-3}\cmidrule(lr){4-5}
$\sigma$ & \NAME & \LIFRC & \NAME & \LIFRC \\
\midrule
0.00 & $69.72\pm0.68$ & $67.15\pm0.66$ & $93.64\pm0.25$ & $67.00\pm1.04$ \\
0.05 & $69.42\pm0.90$ & $67.07\pm0.95$ & $77.22\pm4.07$ & $12.46\pm1.19$ \\
0.10 & $70.10\pm0.81$ & $66.39\pm0.95$ & $73.82\pm3.01$ & $12.16\pm0.63$ \\
0.20 & $68.03\pm0.75$ & $62.67\pm0.80$ & $67.14\pm1.52$ & $12.15\pm0.91$ \\
0.30 & $67.25\pm1.12$ & $57.73\pm0.27$ & $59.90\pm1.34$ & $12.04\pm0.70$ \\
0.50 & $64.09\pm0.97$ & $52.80\pm0.22$ & $46.18\pm1.41$ & $12.08\pm0.69$ \\
\bottomrule
\end{tabular}
\end{table}

\begin{table}[h]
\centering
\caption{Test accuracy (\%) under sampling-rate shifts (rate $=1.0$: original sampling rate; mean~$\pm$~std over three reservoirs). The rate $=1.0$ row reproduces the clean results of Table~\ref{tab:lif_ablation} (five seeds).}
\label{tab:rate}
\small
\begin{tabular}{lcccc}
\toprule
& \multicolumn{2}{c}{\textbf{FordA}} & \multicolumn{2}{c}{\textbf{sMNIST}} \\
\cmidrule(lr){2-3}\cmidrule(lr){4-5}
Rate & \NAME & \LIFRC & \NAME & \LIFRC \\
\midrule
0.70 & $52.02\pm0.55$ & $49.80\pm1.33$ & $7.37\pm2.83$  & $11.22\pm1.41$ \\
0.80 & $53.11\pm3.04$ & $50.66\pm1.60$ & $9.04\pm1.42$  & $13.48\pm2.86$ \\
0.90 & $55.23\pm1.08$ & $53.64\pm3.73$ & $13.06\pm4.20$ & $18.68\pm1.65$ \\
1.00 & $69.72\pm0.68$ & $67.15\pm0.66$ & $93.64\pm0.25$ & $67.00\pm1.04$ \\
1.10 & $57.80\pm1.50$ & $55.68\pm5.84$ & $10.22\pm0.46$ & $25.35\pm8.25$ \\
1.20 & $52.25\pm0.78$ & $53.38\pm4.60$ & $9.80\pm0.00$  & $18.53\pm3.67$ \\
1.30 & $50.66\pm0.37$ & $53.06\pm4.51$ & $9.73\pm0.10$  & $15.69\pm2.44$ \\
\bottomrule
\end{tabular}
\end{table}

\newpage
\input{checklist.tex}

\end{document}

%% file: checklist.tex
\section*{NeurIPS Paper Checklist}



\begin{enumerate}

\item {\bf Claims}
    \item[] Question: Do the main claims made in the abstract and introduction accurately reflect the paper's contributions and scope?
    \item[] Answer: \answerYes{} 
    \item[] Justification: The abstract and introduction state all main contributions and claims, each directly supported by experimental results in Sections 4.1--4.5. 

\item {\bf Limitations}
    \item[] Question: Does the paper discuss the limitations of the work performed by the authors?
    \item[] Answer:  \answerYes{} 
    \item[] Justification: Limitations are discussed in the Conclusion, covering multiple aspects including hardware validation, theoretical characterisation, scope of evaluation, and conditions under which the approach may underperform.

\item {\bf Theory assumptions and proofs}
    \item[] Question: For each theoretical result, does the paper provide the full set of assumptions and a complete (and correct) proof?
    \item[] Answer: \answerNA{} 
    \item[] Justification: The paper does not include theoretical results or proofs; all contributions are empirical.

    \item {\bf Experimental result reproducibility}
    \item[] Question: Does the paper fully disclose all the information needed to reproduce the main experimental results of the paper to the extent that it affects the main claims and/or conclusions of the paper (regardless of whether the code and data are provided or not)?
    \item[] Answer: \answerYes{} 
    \item[] Justification: The architecture and neuron models are fully described in Sections 3.1--3.2, the energy estimation framework in Section 3.3, and the analysis methods in Section 3.4. Full hyperparameter search ranges and selected configurations for all datasets are reported in Appendix A.6 (HRF-Res) and A.5 (LIF-Res baseline), the evaluation and training protocol in Appendix A.4, data preprocessing in Appendix A.3, energy estimation details in Appendix A.2, and frequency sweep and CKA implementation details in Appendices B.1--B.2. Code is publicly available.

\item {\bf Open access to data and code}
    \item[] Question: Does the paper provide open access to the data and code, with sufficient instructions to faithfully reproduce the main experimental results, as described in supplemental material?
    \item[] Answer:\answerYes{} 

    \item[] Justification: Code to reproduce all experiments, tables, and figures is publicly available, together with instructions on environment setup, data download and preprocessing, and the commands to run each experiment. All datasets used are publicly available: MNIST, CIFAR-10, UCR archive, SHD, DVS~Gesture, and N-MNIST. Full implementation details are also provided in Appendices A--C.

\item {\bf Experimental setting/details}
    \item[] Question: Does the paper specify all the training and test details (e.g., data splits, hyperparameters, how they were chosen, type of optimizer) necessary to understand the results?
    \item[] Answer:  \answerYes{} 
    \item[] Justification: All training and evaluation details are provided, including data splits and preprocessing (Appendix A.3), the readout training procedure and optimizer (Appendix A.4), hyperparameter search protocol and selected configurations for all datasets (Appendices A.5--A.6), and compute resources (Appendix A.4). Key details are also summarised in the main text in Section 3.

\item {\bf Experiment statistical significance}
    \item[] Question: Does the paper report error bars suitably and correctly defined or other appropriate information about the statistical significance of the experiments?
    \item[] Answer: \answerYes{} 
    \item[] Justification: All results are reported as mean $\pm$ standard deviation over 5 independent trials, where each trial uses a different random seed controlling all fixed weight matrices and, for neuromorphic datasets, the sparse input projection mask. This is stated consistently across all tables and figures, and described in Appendix A.4.

\item {\bf Experiments compute resources}
    \item[] Question: For each experiment, does the paper provide sufficient information on the computer resources (type of compute workers, memory, time of execution) needed to reproduce the experiments?
    \item[] Answer: \answerYes{} 
    \item[] Justification: Compute resources are reported in Appendix A.4: all experiments were run on a single NVIDIA A100 GPU (40 GB), with per-trial feature extraction ranging from $\sim$10s (sMNIST, $N=800$) to $\sim$15 min (N-MNIST, $N=3600$), and a total compute budget of approximately 500 GPU-hours across all experiments including hyperparameter search. This figure covers all experiments reported in the paper.
    
\item {\bf Code of ethics}
    \item[] Question: Does the research conducted in the paper conform, in every respect, with the NeurIPS Code of Ethics \url{https://neurips.cc/public/EthicsGuidelines}?
    \item[] Answer: \answerYes{} 
    \item[] Justification: The research conforms with the NeurIPS Code of Ethics. The work involves no human subjects, sensitive data, or potential for direct misuse. 

\item {\bf Broader impacts}
    \item[] Question: Does the paper discuss both potential positive societal impacts and negative societal impacts of the work performed?
    \item[] Answer: \answerNA{} 
    \item[] Justification: This work presents foundational research on spiking reservoir computing and is not tied to any specific application or deployment. The primary contribution is a more energy-efficient neural computation model, which has positive implications for low-power edge computing but no direct path to harmful applications.
    
\item {\bf Safeguards}
    \item[] Question: Does the paper describe safeguards that have been put in place for responsible release of data or models that have a high risk for misuse (e.g., pre-trained language models, image generators, or scraped datasets)?
    \item[] Answer: \answerNA{} 
    \item[] Justification: The paper poses no such risks. The model is a small-scale spiking reservoir for time series classification, and all datasets used are standard, publicly available benchmarks with no sensitive content.

\item {\bf Licenses for existing assets}
    \item[] Question: Are the creators or original owners of assets (e.g., code, data, models), used in the paper, properly credited and are the license and terms of use explicitly mentioned and properly respected?
    \item[] Answer: \answerYes{} 
    \item[] Justification: All datasets are properly cited. All are publicly available benchmarks for research use. The software libraries used (PyTorch, scikit-learn) are open-source under BSD-style licenses. Explicit license names are not stated in the paper but all assets are standard research benchmarks with well-established terms of use.

\item {\bf New assets}
    \item[] Question: Are new assets introduced in the paper well documented and is the documentation provided alongside the assets?
    \item[] Answer: \answerNA{} 
    \item[] Justification: No new datasets, pretrained models, or other assets are released with this paper. Code is publicly available.

\item {\bf Crowdsourcing and research with human subjects}
    \item[] Question: For crowdsourcing experiments and research with human subjects, does the paper include the full text of instructions given to participants and screenshots, if applicable, as well as details about compensation (if any)? 
    \item[] Answer: \answerNA{} 
    \item[] Justification: The paper does not involve crowdsourcing or research with human subjects.

\item {\bf Institutional review board (IRB) approvals or equivalent for research with human subjects}
    \item[] Question: Does the paper describe potential risks incurred by study participants, whether such risks were disclosed to the subjects, and whether Institutional Review Board (IRB) approvals (or an equivalent approval/review based on the requirements of your country or institution) were obtained?
    \item[] Answer: \answerNA{} 
    \item[] Justification: The paper does not involve crowdsourcing or research with human subjects, so IRB approval was not required.

\item {\bf Declaration of LLM usage}
    \item[] Question: Does the paper describe the usage of LLMs if it is an important, original, or non-standard component of the core methods in this research? Note that if the LLM is used only for writing, editing, or formatting purposes and does \emph{not} impact the core methodology, scientific rigor, or originality of the research, declaration is not required.
    \item[] Answer: \answerNA{} 
    \item[] Justification: LLMs were used only for writing and editing purposes and did not impact the core methodology, scientific rigor, or originality of the research.

\end{enumerate}

%% file: bibliography.bib
@inproceedings{rusch2021coupled,
  title={Coupled Oscillatory Recurrent Neural Network ({coRNN}): An accurate and (gradient) stable architecture for learning long time dependencies},
  author={Rusch, T. Konstantin and Mishra, Siddhartha},
  booktitle={International Conference on Learning Representations},
  year={2021}
}

@inproceedings{rusch2021unicornn,
  title={{UnICORNN}: A recurrent model for learning very long time dependencies},
  author={Rusch, T. Konstantin and Mishra, Siddhartha},
  booktitle={International Conference on Machine Learning},
  pages={9168--9178},
  year={2021},
  organization={PMLR}
}

@inproceedings{rusch2022graphcon,
  title={Graph-Coupled Oscillator Networks},
  author={Rusch, T. Konstantin and Chamberlain, Benjamin P. and Rowbottom, James and Mishra, Siddhartha and Bronstein, Michael M.},
  booktitle={International Conference on Machine Learning},
  pages={18888--18909},
  year={2022},
  organization={PMLR}
}

@inproceedings{miyato2025akorn,
  title={Artificial {K}uramoto Oscillatory Neurons},
  author={Miyato, Takeru and L{\"o}we, Sindy and Geiger, Andreas and Welling, Max},
  booktitle={International Conference on Learning Representations},
  year={2025}
}

@inproceedings{reichert2014neuronal,
  title={Neuronal Synchrony in Complex-Valued Deep Networks},
  author={Reichert, David P. and Serre, Thomas},
  booktitle={International Conference on Learning Representations},
  year={2014}
}

@article{effenberger2025functional,
  title={The functional role of oscillatory dynamics in neocortical circuits: A computational perspective},
  author={Effenberger, Felix and Carvalho, Pedro and Dubinin, Igor and Singer, Wolf},
  journal={Proceedings of the National Academy of Sciences},
  volume={122},
  number={4},
  pages={e2412830122},
  year={2025}
}

@article{baronig2025advancing,
  title={Advancing spatio-temporal processing through adaptation in spiking neural networks},
  author={Baronig, Michael and Ferrand, Romain and Sabathiel, Silvester and Legenstein, Robert},
  journal={Nature Communications},
  volume={16},
  pages={5776},
  year={2025}
}

@article{hochreiter1997long,
  title={Long short-term memory},
  author={Hochreiter, Sepp and Schmidhuber, J{\"u}rgen},
  journal={Neural computation},
  volume={9},
  number={8},
  pages={1735--1780},
  year={1997},
  publisher={MIT Press}
}

@inproceedings{zhang2025dendritic,
  title={Dendritic Resonate-and-Fire Neuron for Effective and Efficient Long Sequence Modeling},
  author={Zhang, Dehao and Zhang, Malu and Wang, Shuai and Wang, Jingya and Wei, Wenjie and Ma, Zeyu and Wang, Guoqing and Yang, Yang and Li, Haizhou},
  booktitle={Advances in Neural Information Processing Systems},
  volume={38},
  year={2025}
}

@inproceedings{cho2014learning,
    title = "Learning Phrase Representations using {RNN} Encoder{--}Decoder for Statistical Machine Translation",
    author = {Cho, Kyunghyun  and
      van Merri{\"e}nboer, Bart  and
      Gulcehre, Caglar  and
      Bahdanau, Dzmitry  and
      Bougares, Fethi  and
      Schwenk, Holger  and
      Bengio, Yoshua},
    editor = "Moschitti, Alessandro  and
      Pang, Bo  and
      Daelemans, Walter",
    booktitle = "Proceedings of the 2014 Conference on Empirical Methods in Natural Language Processing ({EMNLP})",
    month = oct,
    year = "2014",
    address = "Doha, Qatar",
    publisher = "Association for Computational Linguistics",
    url = "https://aclanthology.org/D14-1179/",
    doi = "10.3115/v1/D14-1179",
    pages = "1724--1734"
}

@inproceedings{vaswani2017attention,
  title={Attention is all you need},
  author={Vaswani, Ashish and Shazeer, Noam and Parmar, Niki and
          Uszkoreit, Jakob and Jones, Llion and Gomez, Aidan N and
          Kaiser, {\L}ukasz and Polosukhin, Illia},
  booktitle={Advances in Neural Information Processing Systems},
  volume={30},
  year={2017}
}

@techreport{jaeger2001echo,
  title={The ``echo state'' approach to analysing and training
         recurrent neural networks},
  author={Jaeger, Herbert},
  institution={German National Research Center for Information Technology},
  number={GMD Report 148},
  year={2001}
}

@article{maass2002real,
  title={Real-time computing without stable states: a new framework
         for neural computation based on perturbations},
  author={Maass, Wolfgang and Natschl{\"a}ger, Thomas and Markram, Henry},
  journal={Neural computation},
  volume={14},
  number={11},
  pages={2531--2560},
  year={2002},
  publisher={MIT Press}
}

@inproceedings{ceni2024ron,
  title={Random Oscillators Network for Time Series Processing},
  author={Ceni, Andrea and Cossu, Andrea and St{\"o}lzle, Maximilian W
          and Liu, Jingyue and Della Santina, Cosimo and Bacciu, Davide
          and Gallicchio, Claudio},
  booktitle={Proceedings of The 27th International Conference on
             Artificial Intelligence and Statistics},
  pages={4807--4815},
  year={2024},
  volume={238},
  series={Proceedings of Machine Learning Research},
  publisher={PMLR}
}

@article{maass1997networks,
  title={Networks of spiking neurons: the third generation of neural
         network models},
  author={Maass, Wolfgang},
  journal={Neural networks},
  volume={10},
  number={9},
  pages={1659--1671},
  year={1997},
  publisher={Elsevier}
}

@article{davies2018loihi,
  title={{Loihi}: A neuromorphic manycore processor with on-chip learning},
  author={Davies, Mike and Srinivasa, Narayan and Lin, Tsung-Han and
          Chinya, Gautham and Cao, Yongqiang and Choday, Sri Harsha and
          Dimou, Georgios and Joshi, Prasad and Knag, Phil and Kulandaiswamy,
          Raghavan and others},
  journal={IEEE Micro},
  volume={38},
  number={1},
  pages={82--99},
  year={2018},
  publisher={IEEE}
}

@article{merolla2014truenorth,
  title={A million spiking-neuron integrated circuit with a scalable
         communication network and interface},
  author={Merolla, Paul A and Arthur, John V and Alvarez-Icaza, Rodrigo
          and Cassidy, Andrew S and Sawada, Jun and Akopyan, Filipp and
          Jackson, Bryan L and Imam, Nabil and Guo, Chen and Nakamura,
          Yutaka and others},
  journal={Science},
  volume={345},
  number={6197},
  pages={668--673},
  year={2014},
  publisher={American Association for the Advancement of Science}
}

@article{izhikevich2001resonate,
  title={Resonate-and-fire neurons},
  author={Izhikevich, Eugene M},
  journal={Neural Networks},
  volume={14},
  number={6-7},
  pages={883--894},
  year={2001},
  publisher={Elsevier}
}

@article{burkitt2006review,
  title={A review of the integrate-and-fire neuron model:
         {I}. Homogeneous synaptic input},
  author={Burkitt, Anthony N},
  journal={Biological cybernetics},
  volume={95},
  number={1},
  pages={1--19},
  year={2006},
  publisher={Springer}
}

@article{llinas1988intrinsic,
  title={The intrinsic electrophysiological properties of mammalian
         neurons: insights into central nervous system function},
  author={Llin{\'a}s, Rodolfo R},
  journal={Science},
  volume={242},
  number={4886},
  pages={1654--1664},
  year={1988},
  publisher={American Association for the Advancement of Science}
}

@article{alonso1989subthreshold,
  title={Subthreshold {Na}$^+$-dependent theta-like rhythmicity in
         stellate cells of entorhinal cortex layer {II}},
  author={Alonso, Angel and Llin{\'a}s, Rodolfo R},
  journal={Nature},
  volume={342},
  number={6246},
  pages={175--177},
  year={1989},
  publisher={Nature Publishing Group}
}

@article{furber2014spinnaker,
  title={The {SpiNNaker} project},
  author={Furber, Steve B and Galluppi, Francesco and Temple, Steve
          and Plana, Luis A},
  journal={Proceedings of the IEEE},
  volume={102},
  number={5},
  pages={652--665},
  year={2014},
  publisher={IEEE}
}

@article{gallego2022eventsurvey,
  title={Event-based vision: A survey},
  author={Gallego, Guillermo and Delbr{\"u}ck, Tobi and Orchard, Garrick
          and Bartolozzi, Chiara and Taba, Brian and Censi, Andrea and
          Leutenegger, Stefan and Davison, Andrew J and Conradt, J{\"o}rg
          and Daniilidis, Kostas and others},
  journal={IEEE Transactions on Pattern Analysis and Machine Intelligence},
  volume={44},
  number={1},
  pages={154--180},
  year={2022},
  publisher={IEEE}
}

@inproceedings{lanthaler2023neural,
  title={Neural Oscillators are Universal},
  author={Lanthaler, Samuel and Rusch, T. Konstantin and Mishra, Siddhartha},
  booktitle={Advances in Neural Information Processing Systems},
  volume={36},
  year={2023}
}

@article{buzsaki2004neuronal,
  title={Neuronal oscillations in cortical networks},
  author={Buzs{\'a}ki, Gy{\"o}rgy and Draguhn, Andreas},
  journal={Science},
  volume={304},
  number={5679},
  pages={1926--1929},
  year={2004},
  publisher={American Association for the Advancement of Science}
}

@incollection{paugam2012computing,
  title={Computing with spiking neuron networks},
  author={Paugam-Moisy, H{\'e}l{\`e}ne and Bohte, Sander},
  booktitle={Handbook of Natural Computing},
  pages={335--376},
  year={2012},
  publisher={Springer},
  editor={Rozenberg, Grzegorz and B{\"a}ck, Thomas and Kok, Joost N.}
}

@book{atkinson1989introduction,
  title={An Introduction to Numerical Analysis},
  author={Atkinson, Kendall E.},
  edition={2nd},
  year={1989},
  publisher={John Wiley \& Sons},
  address={New York}
}

@incollection{lukosevicius2012practical,
  title={A practical guide to applying echo state networks},
  author={Luko\v{s}evi\v{c}ius, Mantas},
  booktitle={Neural Networks: Tricks of the Trade},
  pages={659--686},
  year={2012},
  publisher={Springer}
}

@inproceedings{horowitz2014computing,
  title={1.1 Computing's energy problem (and what we can do about it)},
  author={Horowitz, Mark},
  booktitle={2014 {IEEE} International Solid-State Circuits Conference
             Digest of Technical Papers ({ISSCC})},
  pages={10--14},
  year={2014},
  organization={IEEE},
  doi={10.1109/ISSCC.2014.6757323}
}

@inproceedings{yao2023spike,
  title={Spike-driven Transformer},
  author={Yao, Man and Hu, Jiakui and Zhou, Zhaokun and Yuan, Li and
          Tian, Yonghong and Xu, Bo and Li, Guoqi},
  booktitle={Advances in Neural Information Processing Systems},
  volume={36},
  year={2023}
}

@article{deckers2022elsm,
  title={Extended liquid state machines for speech recognition},
  author={Deckers, Lucas and Tsang, Ing Jyh and Van Leekwijck, Werner
          and Latr{\'e}, Steven},
  journal={Frontiers in Neuroscience},
  volume={16},
  pages={1023470},
  year={2022},
  publisher={Frontiers Media SA},
  doi={10.3389/fnins.2022.1023470}
}

@article{pereznieves2021neural,
  author  = {Perez-Nieves, Nicolas and Leung, Vincent C. H. and Dragotti, Pier Luigi and Goodman, Dan F. M.},
  title   = {Neural heterogeneity promotes robust learning},
  journal = {Nature Communications}, volume = {12}, number = {1}, pages = {5791}, year = {2021}
}

@inproceedings{cristianini2002kernel,
  title={On kernel-target alignment},
  author={Cristianini, Nello and Shawe-Taylor, John and Elisseeff,
          Andr{\'e} and Kandola, Jaz},
  booktitle={Advances in Neural Information Processing Systems},
  volume={14},
  year={2002},
  publisher={MIT Press}
}

@inproceedings{kornblith2019similarity,
  title={Similarity of Neural Network Representations Revisited},
  author={Kornblith, Simon and Norouzi, Mohammad and Lee, Honglak
          and Hinton, Geoffrey},
  booktitle={Proceedings of the 36th International Conference on
             Machine Learning},
  pages={3519--3529},
  series={Proceedings of Machine Learning Research},
  volume={97},
  year={2019},
  publisher={PMLR}
}

@article{dau2019ucr,
  title={The {UCR} time series archive},
  author={Dau, Hoang Anh and Bagnall, Anthony and Kamgar, Kaveh and
          Yeh, Chin-Chia Michael and Zhu, Yan and Gharghabi, Shaghayegh
          and Ratanamahatana, Chotirat Ann and Keogh, Eamonn},
  journal={IEEE/CAA Journal of Automatica Sinica},
  volume={6},
  number={6},
  pages={1293--1305},
  year={2019},
  publisher={IEEE}
}

@article{cramer2020heidelberg,
  title={The {H}eidelberg spiking data sets for the systematic evaluation
         of spiking neural networks},
  author={Cramer, Benjamin and Stradmann, Yannik and Schemmel, Johannes
          and Zenke, Friedemann},
  journal={IEEE Transactions on Neural Networks and Learning Systems},
  volume={33},
  number={7},
  pages={2744--2757},
  year={2022},
  publisher={IEEE}
}

@inproceedings{amir2017dvs,
  title={A low power, fully event-based gesture recognition system},
  author={Amir, Arnon and Taba, Brian and Berg, David and Melano, Timothy
          and McKinstry, Jeffrey and Di Nolfo, Carmelo and Nayak, Tapan
          and Andreopoulos, Alexander and Garreau, Guillaume and
          Mendoza, Marcela and others},
  booktitle={Proceedings of the IEEE Conference on Computer Vision and
             Pattern Recognition},
  pages={7243--7252},
  year={2017}
}

@inproceedings{dey2022efficient,
  title={Efficient Time Series Classification using Spiking Reservoir},
  author={Dey, Sounak and Banerjee, Dighanchal and George, Arun M
          and Mukherjee, Arijit and Pal, Arnab},
  booktitle={2022 International Joint Conference on Neural Networks
             ({IJCNN})},
  pages={1--8},
  year={2022},
  organization={IEEE},
  doi={10.1109/IJCNN55064.2022.9892728}
}

@article{patino2022spinnaker,
  title={Liquid State Machine on {SpiNNaker} for Spatio-Temporal
         Classification Tasks},
  author={Pati\~{n}o-Saucedo, Alberto and Rostro-Gonz\'{a}lez, Horacio
          and Serrano-Gotarredona, Teresa and Linares-Barranco,
          Bernab\'{e}},
  journal={Frontiers in Neuroscience},
  volume={16},
  pages={819063},
  year={2022},
  publisher={Frontiers Media SA},
  doi={10.3389/fnins.2022.819063}
}

@inproceedings{biswas2024icons,
author = {Biswas, Anmol and Medhe, Sharvari Ashok and Singhal, Raghav and Ganguly, Udayan},
title = {Temporal and Spatial Reservoir Ensembling Techniques for Liquid State Machines},
year = {2025},
isbn = {9798350368659},
publisher = {IEEE Press},
url = {https://doi.org/10.1109/ICONS62911.2024.00019},
doi = {10.1109/ICONS62911.2024.00019},
booktitle = {Proceedings of the International Conference on Neuromorphic Systems},
pages = {78–81},
numpages = {4},
location = {Arlington, Virginia, USA},
series = {ICONS '24}
}

@article{gaurav2023reservoir,
  title={Reservoir based spiking models for univariate time series
         classification},
  author={Gaurav, Ramashish and Stewart, Terrence C and Yi, Yang},
  journal={Frontiers in Computational Neuroscience},
  volume={17},
  pages={1148284},
  year={2023},
  publisher={Frontiers Media SA},
  doi={10.3389/fncom.2023.1148284}
}

@article{lecun1998mnist,
  title={Gradient-based learning applied to document recognition},
  author={LeCun, Yann and Bottou, L{\'e}on and Bengio, Yoshua
          and Haffner, Patrick},
  journal={Proceedings of the IEEE},
  volume={86},
  number={11},
  pages={2278--2324},
  year={1998},
  publisher={IEEE}
}

@article{orchard2015nmnist,
  title={Converting static image datasets to spiking neuromorphic
         datasets using saccades},
  author={Orchard, Garrick and Jayawant, Ajinkya and Cohen, Gregory K
          and Thakor, Nitish},
  journal={Frontiers in Neuroscience},
  volume={9},
  pages={437},
  year={2015},
  publisher={Frontiers Media SA},
  doi={10.3389/fnins.2015.00437}
}

@article{lukosevicius2009reservoir,
  title={Reservoir computing approaches to recurrent neural
         network training},
  author={Luko\v{s}evi\v{c}ius, Mantas and Jaeger, Herbert},
  journal={Computer Science Review},
  volume={3},
  number={3},
  pages={127--149},
  year={2009},
  publisher={Elsevier}
}

@inproceedings{higuchi2024balanced,
author = {Higuchi, Saya and Kairat, Sebastian and Boht\'{e}, Sander M. and Otte, Sebastian},
title = {Balanced resonate-and-fire neurons},
year = {2024},
publisher = {JMLR.org},
booktitle = {Proceedings of the 41st International Conference on Machine Learning},
articleno = {735},
numpages = {19},
location = {Vienna, Austria},
series = {ICML'24}
}

@inproceedings{whichmodel,
  author={Grzyb, Beata J. and Chinellato, Eris and Wojcik, Grzegorz M. and Kaminski, Wieslaw A.},
  booktitle={2009 International Joint Conference on Neural Networks}, 
  title={Which model to use for the Liquid State Machine?}, 
  year={2009},
  volume={},
  number={},
  pages={1018-1024},
  doi={10.1109/IJCNN.2009.5178822}}

@article{Frady2022,
   author = {E. Paxon Frady and Sophia Sanborn and Sumit Bam Shrestha and Daniel Ben Dayan Rubin and Garrick Orchard and Friedrich T. Sommer and Mike Davies},
   doi = {10.1007/s11265-022-01772-5},
   issn = {19398115},
   journal = {Journal of Signal Processing Systems},
   volume = {94},
   number = {10},
   pages = {917--927},
   publisher = {Springer},
   title = {Efficient Neuromorphic Signal Processing with Resonator Neurons},
   year = {2022}
}

@inproceedings{Hille2022,
author = {Hille, Julian and Auge, Daniel and Grassmann, Cyprian and Knoll, Alois},
title = {Resonate-and-Fire Neurons for Radar Interference Detection},
year = {2022},
isbn = {9781450397896},
publisher = {Association for Computing Machinery},
address = {New York, NY, USA},
url = {https://doi.org/10.1145/3546790.3546816},
doi = {10.1145/3546790.3546816},
booktitle = {Proceedings of the International Conference on Neuromorphic Systems 2022},
articleno = {23},
numpages = {4},
location = {Knoxville, TN, USA},
series = {ICONS '22}
}

@inproceedings{Shaaban2024,
  author={Shaaban, Ahmed and Chaabouni, Zeineb and Strobel, Maximilian and Furtner, Wolfgang and Weigel, Robert and Lurz, Fabian},
  booktitle={2024 Smart Systems Integration Conference and Exhibition (SSI)}, 
  title={Resonate-and-Fire Spiking Neurons for Target Detection and Hand Gesture Recognition: A Hybrid Approach}, 
  year={2024},
  volume={},
  number={},
  pages={1-6},
  doi={10.1109/SSI63222.2024.10740507}}

@article{AlKhamissi2021,
   author = {Badr AlKhamissi and Muhammad ElNokrashy and David Bernal-Casas},
   journal = {arXiv preprint arXiv:2109.08234},
   title = {Deep Spiking Neural Networks with Resonate-and-Fire Neurons},
   url = {http://arxiv.org/abs/2109.08234},
   year = {2021}
}
